 \documentclass[final,5p,times,twocolumn,authoryear]{elsarticle}

\usepackage[T1]{fontenc}
\usepackage{amssymb,amsmath,amsfonts}
\usepackage[linesnumbered, ruled]{algorithm2e}
\SetAlFnt{\small}
\usepackage{array}
\usepackage[font=normalsize,labelfont=sf,textfont=sf]{subfig}
\usepackage{textcomp}
\usepackage{stfloats}
\usepackage{url}
\usepackage{verbatim}
\usepackage{graphicx}
\usepackage{cite}
\usepackage{multirow}
\usepackage{listings}
\usepackage{xcolor}
\usepackage{caption}
\usepackage{hyperref}
\usepackage{booktabs}
\usepackage[normalem]{ulem} 
\SetCommentSty{mycommfont}

\journal{ISPRS Journal of Photogrammetry and Remote Sensing}

\begin{document}

\begin{frontmatter}



\title{Mind the Modality Gap:\\ Towards a Remote Sensing Vision-Language Model via Cross-modal Alignment}

\author[1,2]{Angelos Zavras}
\author[2]{Dimitrios Michail}
\author[3,4]{Beg\"um Demir}
\author[1]{Ioannis Papoutsis}

\affiliation[1]{organization={Orion Lab, National Observatory of Athens \& National Technical University of Athens}, country={Greece}}
\affiliation[2]{organization={Department of Informatics \& Telematics, Harokopio University of Athens}, country={Greece}}
\affiliation[3]{organization={Faculty of Electrical Engineering and Computer Science, Technische Universit\"at Berlin}, country={Germany}}
\affiliation[4]{organization={BIFOLD - Berlin Institute for the Foundations of Learning and Data}, country={Germany}}

\begin{abstract}
Deep Learning (DL) is undergoing a paradigm shift with the emergence of foundation models. In this work, we focus on Contrastive Language-Image Pre-training (CLIP), a Vision-Language foundation model that achieves high accuracy across various image classification tasks and often rivals fully supervised baselines, despite not being explicitly trained for those tasks. Nevertheless, there are still domains where zero-shot CLIP performance is far from optimal, such as Remote Sensing (RS) and medical imagery. These domains do not only exhibit fundamentally different distributions compared to natural images, but also commonly rely on complementary modalities, beyond RGB, to derive meaningful insights. To this end, we propose a methodology to align distinct RS image modalities with the visual and textual modalities of CLIP. Our two-stage procedure addresses the aforementioned distribution shift, extends the zero-shot capabilities of CLIP and enriches CLIP's shared embedding space with domain-specific knowledge. Initially, we robustly fine-tune CLIP according to the PAINT~\citep{ilharco2022patching} patching protocol, in order to deal with the distribution shift. Building upon this foundation, we facilitate the cross-modal alignment of a RS modality encoder by distilling knowledge from the CLIP visual and textual encoders. We empirically show that both patching and cross-modal alignment translate to significant performance gains, across several RS imagery classification and cross-modal retrieval benchmark datasets. Patching dramatically improves RS imagery (RGB) classification (BigEarthNet-5: +39.76\% mAP, BigEarthNet-19: +56.86\% mAP, BigEarthNet-43: +28.43\% mAP, SEN12MS: +20.61\% mAP, EuroSAT: +5.98\% Acc), while it maintains performance on the representative supported task (ImageNet), and most critically it outperforms existing RS-specialized CLIP variants such as RemoteCLIP~\citep{liu2023remoteclip} and SkyCLIP~\citep{wang2024skyscript}. Cross-modal alignment extends zero-shot capabilities to multi-spectral data, surpassing our patched CLIP classification performance and establishing strong cross-modal retrieval baselines. Linear probing further confirms the quality of learned representations of our aligned multi-spectral encoder, outperforming existing RS foundation models such as SatMAE~\citep{cong2022satmae}. Notably, these enhancements are achieved without the reliance on textual descriptions, without introducing any task-specific parameters, without training from scratch and without catastrophic forgetting. Our work highlights the potential of leveraging existing VLMs' large-scale pre-training and extending their zero-shot capabilities to specialized fields, paving the way for resource efficient establishment of in-domain multi-modal foundation models in RS and beyond. We make our code implementation and weights for all experiments publicly available on our project's GitHub repository \url{https://github.com/Orion-AI-Lab/MindTheModalityGap}.
\end{abstract}


\end{frontmatter}



\section{Introduction}
The Earth's orbit is swarming with a growing constellation of Earth Observation (EO) satellites, continuously producing an unprecedented volume of diverse and complex information about our planet. Notably, as of today, Europe's Copernicus satellite constellation alone is producing more than 20TB of satellite data every day, which as a matter of fact outstrips our capacity to extract meaningful information. Building upon this wealth of data, the field of Remote Sensing (RS) has witnessed a surge in the application of Deep Learning (DL) models, aiming to revolutionize our ability to monitor, understand, and respond to changes on our planet. 

Despite these advances, the performance of such DL models is often constrained by the annotated data scarcity, due to the domain expertise and subsequently the cost and time required for the interpretation of RS images. As a result, these models are exceptionally fragile, designed to excel only within the narrow classification space for which they were specifically trained. This is an important issue for the RS community since various nomenclatures and classification schemes are used to categorize Earth's features. These schemes play a crucial role in classifying and interpreting RS data, enabling effective analysis and information extraction, depending on the objectives of the study, available data, and the desired level of detail and accuracy required for the analysis.

In contrast to typical image classifiers, open-vocabulary models are not constrained to a fixed classification space and are able to perform any image classification task, using textual descriptions of the class names. One the other hand, foundation models~\citep{bommasani2021opportunities}, commonly trained via self-supervision at scale, on large amounts of unlabeled data obtained through web-crawling approaches, are large and versatile deep learning models that can be adapted to a wide range of downstream tasks. Open-vocabulary foundation models, represent a powerful combination of large-scale pre-training and the ability to handle words outside of a fixed vocabulary. These models have gathered considerable attention due to their exceptional performance and ability to generalize across different domains and are particularly useful in scenarios where the textual input may include domain-specific terms.

In this work, we investigate OpenAI's Contrastive Language-Image Pre-Training (CLIP) model, one of the so-called open-vocabulary foundation models, which is an innovative approach to computer vision that leverages natural language supervision at scale. The model architecture consists of two encoders - one for images and one for text - that are trained together to maximize the cosine similarity between paired images and captions in a shared embedding space, while minimizing it for unrelated pairs. CLIP has been trained on a large dataset of noisy image-text pairs collected from the internet, learning to associate images with their textual descriptions. This training paradigm allows CLIP to develop a flexible understanding of visual concepts and perform zero-shot classification and retrieval tasks across a wide range of domains, that can be directed through natural language prompts, without the need for task-specific fine-tuning. For the RS domain, CLIP offers intriguing possibilities. While not specifically trained on RS data, CLIP has presumably been exposed to a marginal quantity of EO-related samples~\citep{czerkawski2023laion}, potentially providing a strong foundation for downstream RS tasks. 
Its zero-shot capabilities enable it to classify images into arbitrary text-specified categories without requiring retraining, making it a promising tool for more flexible and adaptable Earth Observation systems.

While CLIP's versatility and zero-shot capabilities offer intriguing possibilities for various fields, its performance is not uniformly optimal across all domains. In particular, there are specialized domains such as RS and medical imagery, where CLIP's zero-shot performance falls short of domain-specific models. Satellite scene classification, in particular, is one of the few tasks presented by the CLIP authors, where zero-shot CLIP significantly under-performs on the EuroSAT dataset, resulting in the biggest delta (37,1\%) among all tasks assessed, when compared to a fully supervised ResNet50 baseline model. To this end, we identify three major gaps for the task of satellite scene classification which we address in this work. The first gap regards the distribution shift~\citep{koh2021wilds}, i.e. a change in the underlying data distribution that can occur when the data used during training, significantly differs from the data the model encounters when deployed, leading to significant performance degradation. Satellite imagery does not only demonstrate a fundamentally different distribution compared to natural images, but it also poses unique challenges in terms of spatial resolution, atmospheric conditions, and varying perspectives due to different orbits, which can significantly impact the performance of computer vision models. The second gap arises from the information constraint induced by relying solely on the RGB modality. Satellite imagery commonly exploit complementary modalities, beyond RGB, to derive meaningful insights. These modalities include multi-spectral, hyper-spectral and radar data, as well as byproducts of those modalities such as InSAR~\citep{bountos2022hephaestus} data. Leveraging these diverse modalities grants us with critical yet complementary information that contribute to a deeper understanding of our planet. The third gap, pertains to the scarcity of datasets containing pairs of satellite imagery and corresponding textual descriptions. Typically used RS image-text paired datasets~\citep{qu2016deep, lu2017exploring, yuan2022exploring, zhan2023rsvg}, are limited to aerial and Very High-Resolution (VHR) commercial satellite imagery. There are three interrelated drawbacks in these datasets. First, they do not rely on free and open-access data, such as the Copernicus Sentinel satellite constellation data. Second, the spatial resolution of these datasets, which directly relates to the level of detail that can be retrieved from a scene, leads to a different underlying data distribution compared to non-commercial satellite data. Last but not least, they typically incorporate only RGB imagery which limits their spectral information and ignores the valuable multi-spectral and hyper-spectral capabilities of many RS satellites. All three factors limit their usability and effectiveness in downstream RS applications.

In order to address these gaps, we propose a novel methodology comprised of two consecutive stages, working towards cross-modal alignment of RS imagery modalities in the context of CLIP. Our study is based on OpenAI's CLIP collection of pre-trained image-text models. For the first step, we define a patching process to robustly fine-tune CLIP using RS data RGB composites, in order to deal with the aforementioned distribution shift without impacting CLIP zero-shot performance on natural image classification tasks. PAINT (Patching with Interpolation)~\citep{ilharco2022patching} is a method for adapting pre-trained large-scale open-vocabulary models to tasks where the model performs poorly (patching tasks), without compromising tasks where their performance is already adequate (supported tasks). This method is particularly relevant for Earth observation applications, where models like CLIP, pre-trained on diverse image-text pairs, can be fine-tuned for specialized tasks such as LULC scene classification using satellite imagery. PAINT operates by first fine-tuning the model's visual encoder on the patching task using the output of CLIP’s text encoder as a frozen classification layer (frozen classification head) to map image features to the space of classes instead of introducing a learnable classification layer, and then linearly interpolating between the original zero-shot and fine-tuned visual encoder's weights. This interpolation is controlled by a mixing coefficient, optimized to balance the performance trade-off between the patching and the supported tasks. For the RS community, PAINT offers several advantages. It significantly improves accuracy on specific Earth observation tasks while maintaining broad applicability to other vision tasks. This is crucial for developing versatile models that can handle the diverse challenges in satellite and aerial image analysis. Furthermore, PAINT's effectiveness increases with model size, aligning well with the trend towards larger, more powerful models in RS. The method's ability to facilitate "broad transfer", thus improving performance on related tasks even with disjoint classes, is particularly valuable in the context of multi-sensor and multi-scale Earth observation data, where adaptability across varying imagery types and spatial resolutions is essential. As for the second step, we extend CLIP's zero-shot capabilities by distilling knowledge from the CLIP visual and textual encoders, in order to facilitate the cross-modal alignment of a pre-trained RS encoder with the visual and textual modalities of CLIP. This process extends the zero-shot capabilities of CLIP and enriches CLIP's shared embedding space with domain-specific knowledge. We ultimately demonstrate our method on the task of RS imagery classification. We empirically show that both robust fine-tuning and cross-modal alignment translate to notable performance gains, over several RS imagery classification benchmark datasets.

Our main contributions can be summarized as follows:
\begin{itemize}
    \item We propose a novel method for the cross-modal alignment of RS imagery modalities in the context of CLIP, without the reliance on textual descriptions, without introducing any task-specific parameters, without training from scratch and without catastrophic forgetting. Unlike previous approaches in RS domain, our method uniquely harnesses CLIP's large-scale pre-training while adapting it to RS-specific challenges. Crucially, we achieve this adaptation without compromising CLIP's performance in other domains or altering its fundamental shared embedding space characteristics. This approach allows us to benefit from CLIP's broad knowledge base and zero-shot capabilities while tailoring it to the problem at hand, offering a more versatile and powerful tool for RS tasks.
    \item We evaluate the generalization of both patched and aligned models, and provide an extensive benchmark on a series of prominent RS imagery datasets for the task of RS image classification and cross-modal retrieval. In addition, we compare our results with related works in the RS domain that adapt CLIP to the problem at hand. Additionally, we assess the quality of our model's representations through linear probing, training only the last linear layer while using various RS foundation models as the backbone. This extensive evaluation not only demonstrates the effectiveness of our method but also establishes new performance baselines for these tasks in the RS domain.
    \item We set the baseline for RS image classification and cross-modal retrieval tasks, in comparison to a representative CLIP supported task (i.e. ImageNet) proving that our final Vision-Language model (VLM) not only benefits from but also extends CLIP's large-scale pre-training.   
    \item We introduce an enhanced CLIP model that now attains significantly improved zero-shot performance on RS imagery classification and cross-modal retrieval tasks. This enhancement is achieved by effectively capturing the semantic content of RS, overcoming nomenclature limitations and leveraging complementary information, beyond RGB, utilizing available RS imagery modalities. This comprehensive approach allows the model to leverage a wider range of available information, resulting in more robust and accurate performance across various RS applications. We make our code implementation and weights for all experiments publicly available on our project's GitHub repository\footnote{\url{https://github.com/Orion-AI-Lab/MindTheModalityGap}}, encouraging the community to explore, adapt, and extend our implementation to address diverse challenges in RS, aiming to facilitate the development of novel applications based on RS data.
\end{itemize}

\section{Related Work}
\label{sec:related_work}

\subsection{Contrastive Language-Image Pre-training}
Recent developments in the multi-modal contrastive learning era have significantly enhanced the capabilities of CLIP-like models. These models are known for their ability to effectively extract and distinguish between rich visual and textual features~\citep{shen2021much, barraco2022unreasonable, li2023decap} and are suitable for a variety of downstream tasks, such as image classification~\citep{radford2021learning}, action recognition~\citep{wang2021actionclip}, semantic segmentation~\citep{wang2022cris}, text-guided image generation~\citep{patashnik2021styleclip}, image and video captioning~\citep{cornia2021universal, mokady2021clipcap, tang2021clip4caption}. Lately, such models have been also utilized either in part or in their entirety, as versatile building blocks for the next generation of foundation models~\citep{ramesh2021zero, ramesh2022hierarchical, alayrac2022flamingo, kirillov2023segment}. It is evident that CLIP due to its large scale pre-training, as well as its proven efficacy on a wide range of downstream tasks and applications, emerges as an advantageous frontier in the ongoing quest towards foundation models for EO. 

LAION-AI with OpenCLIP~\citep{ilharco_gabriel_2021_5143773}, the open-source implementation of OpenAI's CLIP~\citep{radford2021learning}, has demonstrated impressive results. They managed to replicate OpenAI's proprietary pre-training dataset~\citep{schuhmann2022laion} and subsequently trained and published several models~\citep{cherti2023reproducible}, using various architectures, on a variety of data sources and compute budgets, ranging from small to large-scale experiments. Recent advancements in the context of CLIP pre-training have showcased remarkable achievements along the axes of pre-train data filtering~\citep{gadre2023datacomp, fang2023data, xu2023demystifying}, model architecture~\citep{sun2023eva} and computational efficiency~\citep{li2023inverse, li2023clipa, zhai2023sigmoid}, leading to substantial improvements and eventually establishing new standards within the era of pre-trained CLIP models. These advancements, formulate an imperative demand for an extensive set of benchmarking experiments, aiming to assess the ability of those newly released pre-trained CLIP models to facilitate novel applications, in the context of RS.

\subsection{Domain Specialized CLIP Models}
Data is regarded as the cornerstone for foundation model training and CLIP is no exception. Existing CLIP models are trained on large amounts of web-crawled data, to encode general knowledge, without emphasizing on specialized domains such as RS and medical imagery, which as a matter of fact demonstrate fundamentally different distributions compared to natural images encountered during pre-training.

Several efforts have been made to overcome this phenomenon and adapt CLIP aiming to yield a robust foundation model with proficient expertise in specialized domains. Unlike natural image-text datasets, which can easily reach billion-scale~\citep{schuhmann2022laion, gadre2023datacomp}, specialized datasets hold a relatively limited scale due to their innate demand for highly targeted knowledge. The medical imaging community, for instance, has witnessed a significant surge in contributions of CLIP-centered studies~\citep{wang2022medclip, eslami2023pubmedclip, zhang2023large, zhao2023clip}. They commonly encompass biomedical image-caption pairs, collected and filtered from open-access research papers~\citep{pelka2018radiology, subramanian2020medicat, lin2023pmc, liu2023qilin}, medical reports~\citep{johnson2019mimic, bustos2020padchest, li2021ffa} and social media platforms~\citep{huang2023visual,ikezogwo2023quilt}. On the contrary, the RS domain lags behind, in terms of CLIP-oriented developments, primarily due to the in-domain image-text paired data scarcity. To this end, recent developments predominantly revolve around the exploitation of existing limited extent RS datasets and the data efficient adaptation of CLIP models to the problem at hand. 

\citet{arutiunian2021fine} fine-tuned CLIP by leveraging three small existing RS image captioning datasets and demonstrated their results on retrieval-related tasks. \citet{czerkawski2023detecting} highlighted that in zero-shot setting, CLIP exhibits difficulty detecting cloud-free images and mitigated this limitation through a cost-effective training stage consisting of a few hundred optimization steps of a single linear layer on top of CLIP image encodings, demonstrating improved performance and transferability across various sensor types and spectral bands. \citet{singha2023applenet} proposed APPLeNet, an image-conditioned prompt learning strategy for few-shot RS image generalization using CLIP models. Their approach focuses on multi-scale feature learning and disentangles visual style and content primitives for domain generalization tasks in RS, outperforming zero-shot CLIP in several RS benchmark datasets. \citet{liu2023remoteclip} presented RemoteCLIP, a RS domain specialized CLIP model, fine-tuned using existing RS image captioning datasets, extended using heterogeneous satellite and UAV imagery annotations (e.g bounding boxes and masks) after turning them into image-caption pairs, achieving remarkable results across various datasets and downstream tasks. \citet{zhang2023rs5m} presented RS5M, a 5 million RS image captioning dataset, obtained by filtering publicly available image-text paired datasets and captioning label-only RS datasets with pre-trained VLM, with the purpose of fine-tuning CLIP. They experimented with full fine-tuning and several Parameter-Efficient Fine-Tuning methods and ultimately demonstrated their final model GeoRSCLIP, on zero-shot classification, cross-modal image-text retrieval and semantic localization tasks in comparison to state-of-the-art RS-tailored CLIP models. \citet{wang2024skyscript} introduced SkyScript, a large-scale vision-language dataset for RS with 2.6 million image-text pairs, by connecting open satellite imagery with OpenStreetMap semantic tags. Moreover they obtained a RS subset of LAION-2B~\citep{schuhmann2022laion} dataset (LAION-RS), by applying a binary classification model. They used both datasets to develop two distinct RS-specialized CLIP models, SkyCLIP and CLIP-LAION-RS, by continual pre-training CLIP and evaluating the impact. \citet{yuan2023parameter} introduced a Parameter-Efficient Transfer Learning (PETL)~\citep{houlsby2019parameter} method for RS Image-Text Retrieval, leveraging a pre-trained CLIP model along a multi-modal adapter and a Hybrid Multi-modal Contrastive learning objective, outperforming traditional methods and showcasing a drastic reduction in training costs compared to full fine-tuning. \citet{mo2023s} proposed S-CLIP, a semi-supervised learning method for fine-tuning CLIP that utilizes additional unpaired images by employing pseudo-labeling strategies specifically designed for contrastive learning, significantly enhancing fine-tuning results using fewer image-text pairs than typically required. \citet{bhattacharya2023c} proposed C-SAW, a self-supervised prompt learning technique which incorporates a reconstruction task for better image generalization in RS applications. During fine-tuning, they kept the CLIP backbone frozen and introduced a small set of projectors for both CLIP encoders to train contrastively using C-SAW. \citet{dhakal2023sat2cap} introduced a novel weakly supervised method for creating maps based on free-form textual descriptions, termed zero-shot mapping. They utilized a contrastive learning framework named Sat2Cap, trained on paired overhead and ground-level images, to predict CLIP embeddings of ground-level scenery from satellite images. They managed to map a wide variety of attributes without text-labeled data, overcoming the limitations of previous models that could only map pre-defined attributes. Similarly, \citet{mall2023remote} addressed the scarcity of textual descriptions, by training contrastively an image encoder that maps RGB satellite imagery to the same representation space of a frozen CLIP image encoder, using a large amount of paired internet images and RGB satellite composites. They managed to yield state-of-the-art results across numerous downstream tasks and indicated that leveraging a large number of ground-satellite image pairs without accompanying text is more beneficial than fine-tuning on small datasets, in a supervised manner.  

\setlength{\parskip}{0pt} It is evident that the main focus in tailoring CLIP for RS revolves around overcoming the scarcity of in-domain image-text paired data. Efforts span from leveraging limited-scale and synthetic datasets for fine-tuning to innovative data-efficient and self-supervised strategies, all aimed at optimizing CLIP for RS. Our work stands out by exploiting existing labeled RS imagery data to deal with the RS domain distribution shift, without introducing any task-specific parameters to the pre-trained CLIP models and without catastrophic forgetting.

\subsection{Multi-modal CLIP-inspired models in Remote Sensing}
Multi-modal learning in the RS domain~\citep{rolf2024mission} has evolved significantly, marking a transition from traditional single-source data analysis to the integration of diverse RS data types, such as optical imagery and radar, to attain a more accurate and holistic understanding of the Earth. Over time, advancements in machine learning and data fusion~\citep{li2022deep} techniques have led to sophisticated models that can handle the complexity and variability of multi-modal RS data~\citep{zhang2018change, uzkent2019learning, toker2021coming}. Recently, the introduction of CLIP~\citep{radford2021learning} has significantly influenced the development of RS models capable of seamlessly integrating various modalities, demonstrating an intuitive and versatile approach that extends well beyond the initial image-text framework. This advancement has opened up new possibilities for more intuitive and semantically rich analyses of RS imagery, allowing models to complement and correlate the same Earth features as being observed by different sensors, demonstrating the growing potential of multi-modal learning in capturing the intricate features of our planet.

In this context, \citet{allen2023fewshot} pre-trained a ViT-based CLIP model using three different RS imagery modalities (Sentinel-2 RGB optical and Sentinel-1 SAR radar amplitude and interferometric coherence), across five AOIs covering a small percentage of Earth’s total landmass. The model consists of three separate single channel ViT-based input encoders, operating on whatever channel was randomly selected for each modality, during pre-training, aiming to create a shared embedding space between all three modalities, while similarity is measured for each pair of modalities and then averaged. \citet{klemmer2023satclip} demonstrated SatCLIP, a global, geographic location encoder that learns general-purpose, implicit representations using globally available satellite imagery, pre-trained by matching satellite images and their respective coordinates using the CLIP objective. Along the same lines, \citet{cepeda2023geoclip} introduced GeoCLIP, a CLIP-inspired Image-to-GPS retrieval approach for worldwide geo-localization. Unlike SatCLIP, GeoCLIP leverages a pre-trained CLIP image encoder and enforces alignment with the corresponding GPS locations. Using hierarchical learning and random Fourier features, GeoCLIP demonstrated impressive effectiveness in limited-data settings and potential application in a plethora of downstream tasks. \citet{khanal2023learning} proposed GeoCLAP, a contrastive-learning framework for the task of soundscape mapping based on the relationship between sound and the visual characteristics of geographic locations. They leveraged a CLIP-based model to encode three types of data: geo-tagged audio recordings, textual descriptions of audio, and overhead images of their capture locations. As a result they enabled a unified embedding space, for all three modalities, that can predict the most probable sounds at any given geographic location, outperforming previous state-of-the-art models. 

A fundamental notion that distinguishes our work from related research, is the cross-modal alignment of complementary RS imagery modalities, beyond RGB, within the shared embedding space of CLIP, leveraging its large-scale pre-training. Our endeavors focus on aligning a multi-spectral Sentinel-2 encoder with the RGB-image and text encoders of CLIP, providing a blueprint for the cross-modal alignment of more RS imagery modalities and ultimately enabling a rich set of cross-modal retrieval and text-based zero-shot downstream tasks.

\section{Method}
Our proposed approach addresses the challenge of adapting pre-trained VLMs, such as CLIP (Contrastive Language-Image Pre-training), to the domain of RS while maintaining their broad applicability. This method consists of two main stages: first, we adapt CLIP to satellite imagery using PAINT (Patching with Interpolation) in order to deal with the underlying distribution shift, and second, we align RS modalities with natural images and text in a shared embedding space by distilling knowledge from the CLIP visual and textual encoders. This two-stage process enables multi-modal downstream tasks by facilitating the association between RS imagery modalities, RGB images, and text through learned representations.

The core of our method lies in creating a shared embedding space that effectively aligns different modalities. Let $\mathcal{D} = \{(\mathbf{x}_i, y_i)\}_{i=1}^N$ be a labeled RS imagery dataset consisting of $N$ image-label pairs, where $\mathbf{x}_i$ is the $i$-th RS image and $y_i$ is its corresponding label. Each RS image $\mathbf{x}_i$ has a corresponding RGB composite image $\mathbf{\tilde{x}}_i$, produced using a dedicated function $C_\text{RGB}(\cdot)$ depending on the RS modality of $\mathbf{x}_i$. Given the set of RS images $\mathcal{X} = \{\mathbf{x}_i\}_{i=1}^N$, the corresponding set of RGB composites $\mathcal{\tilde{X}} = \{\mathbf{\tilde{x}}_i\}_{i=1}^N$, and the associated labels $\mathcal{Y} = \{y_i\}_{i=1}^N$, our goal is to learn a shared embedding space $\mathcal{E}$ by utilizing RGB composites and labels as anchors to effectively align these modalities.

We aim to find a mapping function $f: \mathcal{X} \cup \mathcal{\tilde{X}} \cup \mathcal{Y} \rightarrow \mathcal{E}$, such that for each triplet $(\mathbf{x}_i, \mathbf{\tilde{x}}_i, y_i)$, the embeddings $f(\mathbf{x}_i)$, $f(\mathbf{\tilde{x}}_i)$, and $f(y_i)$ are closely aligned in $\mathcal{E}$. The final embedding space $\mathcal{E}$ enables multi-modal downstream tasks by facilitating the association between RS imagery modalities, RGB images, and text through learned representations. To achieve this, we introduce a two-stage approach, explicitly defined in Algorithm~\ref{alg:alignment}.

\subsection{First Stage: Patching CLIP by interpolating weights}
Large pre-trained models such as CLIP~\citep{radford2021learning}, ALIGN~\citep{jia2021scaling} and BASIC~\citep{pham2023combined} have demonstrated unprecedented robustness to a plethora of challenging distribution shifts. However, there are still settings where their zero-shot performance is far from optimal. To this end, Patching with Interpolation (PAINT)~\citep{ilharco2022patching} has emerged as a method that substantially improves accuracy under distribution shift, while maintaining high performance on the target distribution. PAINT employs a two-step procedure, which consists of fine-tuning the model's image encoder via supervised learning using a representative classification task dataset and then linearly interpolating between the weights of the model before and after fine-tuning. This approach enables the expansion of the set of tasks on which models achieve high accuracy, without introducing any task-specific parameters, without re-training them from scratch, and without catastrophic forgetting~\citep{kirkpatrick2017overcoming}.

\begin{figure*}[tbh!]
    \centering
    \includegraphics[width=1.0\linewidth]{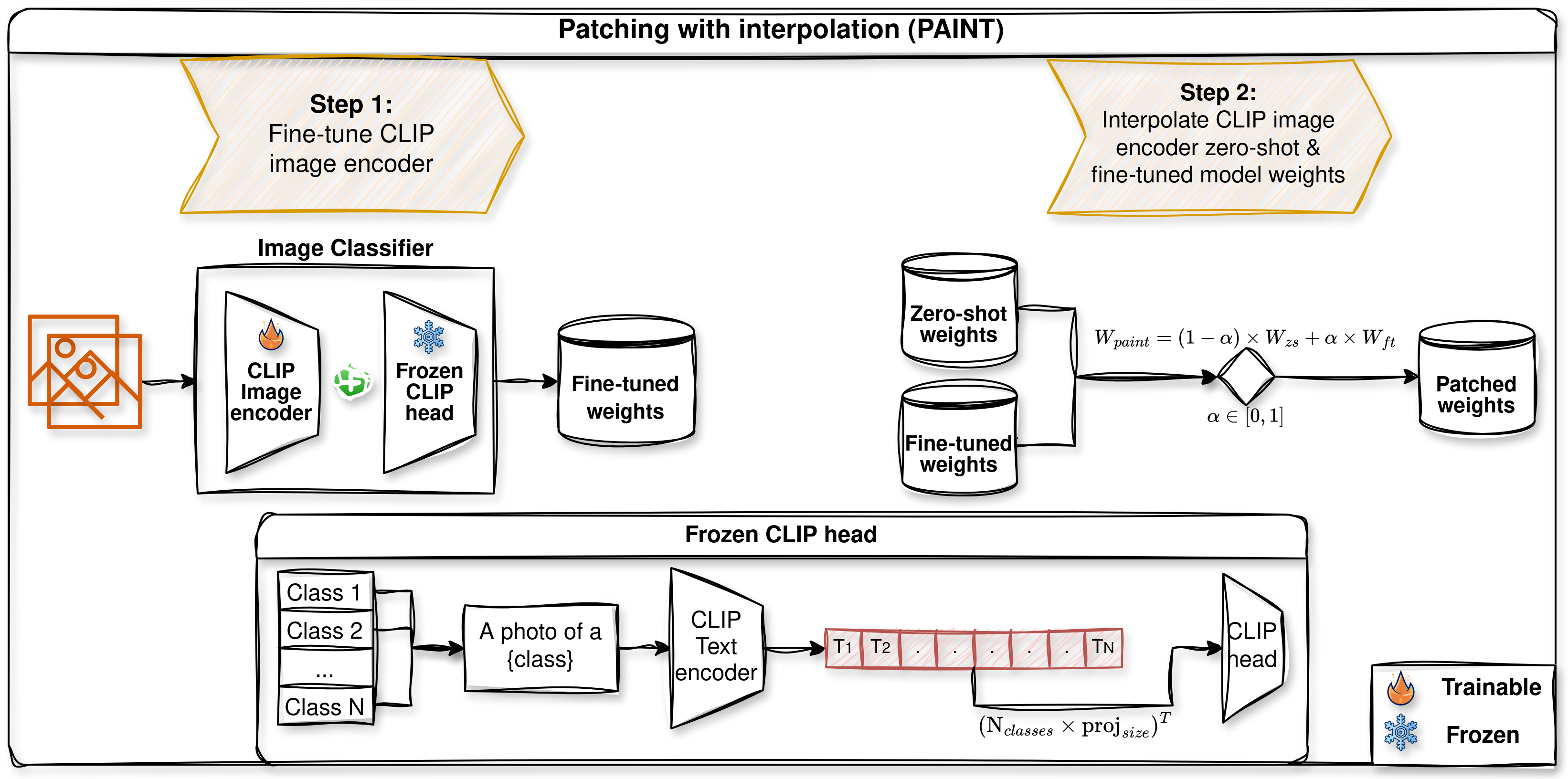}
    \caption{For the first stage of out method we leverage Patching with interpolation (PAINT)~\citep{ilharco2022patching}, in order to deal with the underlying distribution shift of the satellite data, without compromising the performance of CLIP on tasks where performance is already adequate (supported tasks). The PAINT two-step process involves (1) fine-tuning the CLIP image encoder using the output of CLIP’s text encoder as a frozen classification head to map image features to the space of classes, and (2) interpolating between the original zero-shot $W_{zs}$ and fine-tuned $W_{ft}$ model weights using a linear mixing coefficient $\alpha \in [0, 1]$ ending up with the patched weights $W_{paint}=(1-\alpha) \times W_{zs} + \alpha \times W_{ft}$.}
    \label{fig:patching_clip}
\end{figure*}

In order to deal with the distribution shift between natural images and satellite imagery we initially robust fine-tune CLIP, as shown in Fig.~\ref{fig:patching_clip}, following the patching protocol described in \citet{ilharco2022patching}. Eventually, we end up with a refined embedding space, thus providing a solid foundation to facilitate the satellite cross-modal alignment stage. Given an open-vocabulary model, i.e. CLIP, we fine-tune the model's image encoder zero-shot weights $W_{zs}$ on training data from a patching task $D_{patch}$, i.e. a representative classification task dataset where the zero-shot performance is far from optimal, producing fine-tuned weights $W_{ft}$. During fine-tuning, the CLIP image encoder is trained on remote sensing (RS) imagery, using cross-entropy loss to minimize the difference between predicted and actual classification labels. We seek to produce patched weights $W_{patch}$ that achieve high accuracy on $D_{patch}$, without decreasing the performance on supported tasks $D_{supp}$ where performance is already adequate, by linearly interpolating $W_{zs}$ and $W_{ft}$ weights. During the course of fine-tuning, instead of introducing a learnable classification layer, \citet{ilharco2022patching} use the frozen output of CLIP's text encoder, as the output layer of the image encoder in order to map image features to the space of classes. Ultimately, a mixing coefficient $\alpha \in[0,1]$ is determined via held-out validation sets for $D_{supp}$ and $D_{patch}$, in order to linearly interpolate $W_{zs}$ and $W_{ft}$ and produce patched weights $W_{patch} = (1-\alpha) \times W_{zs} + \alpha \times W_{ft}$, where the zero-shot performance is adequate for both supported tasks $D_{supp}$ and patching tasks $D_{patch}$.

\begin{algorithm}[tbh!]
    \SetKwInput{KwInput}{Input}
    \SetKwInput{KwOutput}{Output}
    \SetAlgoLined
    \SetNoFillComment
    \DontPrintSemicolon
    
    \KwInput{Pre-trained CLIP model $\mathcal{M}$ and tokenizer $\mathcal{T}$}
    \KwInput{Dataset $D = \{(x_i,y_i)\}^n_{i=1}, D_{classnames}, D_{prompts}\}$}
    \KwInput{Pre-trained satellite modality encoder $\mathcal{M}_{\text{sat}}$}
    \BlankLine
    \KwOutput{Aligned satellite encoder $\mathcal{M}_{\text{sat}}$}
    \BlankLine
  
    \SetKwFunction{FMain}{main}
    \SetKwFunction{FHead}{cls\_head}
    \SetKwFunction{FAlign}{align}

    \SetKwProg{Fn}{Function}{:}{}
    \tcc{Create classification head.}
    \Fn{\FHead{}}{
        set $\mathcal{M}$ in inference mode\;
        $W_{cls} \gets []$
        
        \For{$cls \in D_{classnames}$}{
            $T \gets []$
            
            \For{$p \in D_{prompts}$}{
                $T \gets T \cup \{ p.format(cls)$\}\;
            }
            $T \gets \mathcal{T}(T)$
            \tcp*[l]{tokenize}
            $\widetilde{E} \gets \mathcal{M}(T)$
            \tcp*[l]{encode text}
            Normalize \& average text embeddings $\widetilde{E}$\;
            $W_{cls} \gets W_{cls} \cup \widetilde{E}$\;
        }
    
        Stack $W_{cls}$, transpose and apply logit scaling\;
        $h \gets$ linear $classification\_head$ using $W_{cls}$\;
       \Return{$h$}
    }
    \;
  
    \tcc{Align models $\mathcal{M}_{\text{teach}}$ and $\mathcal{M}_{\text{stud}}$.}
    \SetKwProg{Fn}{Function}{:}{}
    \Fn{\FAlign{$\mathcal{M}_{\text{teach}}$, $\mathcal{M}_{\text{stud}}$}}{
    initialize frozen classification head $h \gets cls\_head()$\;
    set $\mathcal{M}_{\text{stud}}$ to trainable and freeze $\mathcal{M}_{\text{teach}}$\;
    \For{$\{ x_i, \widetilde{x}_i \}, y_i \in D$}{
        let $x_i, \widetilde{x}_i$ be the two modalities for sample $i$\;
        $E_i \gets \mathcal{M}_{\text{teach}}(x_i)$\;
        $\widetilde{E}_i  \gets \mathcal{M}_{\text{stud}}(\widetilde{x_i})$\;
        $y_{pred} \gets h(\widetilde{E}_i)$\;
        $loss \gets MSE(E_i, \widetilde{E}_{i}) + \lambda \times CE(y_{pred}, y_{i})$\;
    \dots\;
    }
    \Return{$\mathcal{M}_{\text{stud}}$}\;
    }
    \;
    
    \SetKwProg{Fn}{Function}{:}{}
    \Fn{\FMain{}}{
    \tcc{Step 1: CLIP model Patching}
    initialize CLIP model $\mathcal{M}$ and tokenizer $\mathcal{T}$\;
    $clip\_visual \gets$ image encoder of $\mathcal{M}$\;
    $clip\_visual.fc \gets cls\_head()$\;    
    \BlankLine
    $W_{ft} \gets$ finetune $clip\_visual$ weights $W_{zs}$ on dataset $D$\;
    $W_{paint} \gets (1 - \alpha) \times W_{zs} + \alpha \times W_{ft}$\;
    
    \BlankLine
    \tcc{Step 2: Cross-modal alignment}
    initialize satellite modality encoder $\mathcal{M}_{\text{sat}}$\;
    $clip\_visual \gets clip\_visual.load(W_{paint})$\;
    $\mathcal{M}_{\text{sat}} \gets align(clip\_visual,\mathcal{M}_{\text{sat}},D)$\;
    
    \Return{$\mathcal{M}_{\text{sat}}$}\;
}
\caption{Cross-modal alignment method.}
\label{alg:alignment}
\end{algorithm}

\subsection{Second Stage: Aligning RS imagery modalities with natural images and text}
The main objective of contrastive representation learning is to create a shared embedding space where similar sample pairs are close together and dissimilar pairs are far apart. This technique can be applied in both supervised and unsupervised learning scenarios, with particular effectiveness in self-supervised learning. Contrastive learning also allows for aligning specific pairs of modalities, such as images and text or audio and text. By using pairs of related and unrelated examples, contrastive learning enables the alignment of these modalities in a joint embedding space. However, it is important to note that the resulting embeddings are specific to the aligned modalities and may not directly transfer to other modalities. For instance, embeddings obtained from aligning image and audio modalities may not be directly applicable to text-based tasks. CLIP leverages the aligned (image-text) embedding space for the zero-shot classification task. Given a list of textual descriptions generated using the candidate classes, an input image is classified based on their similarity in the embedding space. Extending the zero-shot classification to other modalities usually requires additional training using paired text data (e.g., audio, text).

\begin{figure*}[tbh!]
    \centering
    \includegraphics[width=1\linewidth]{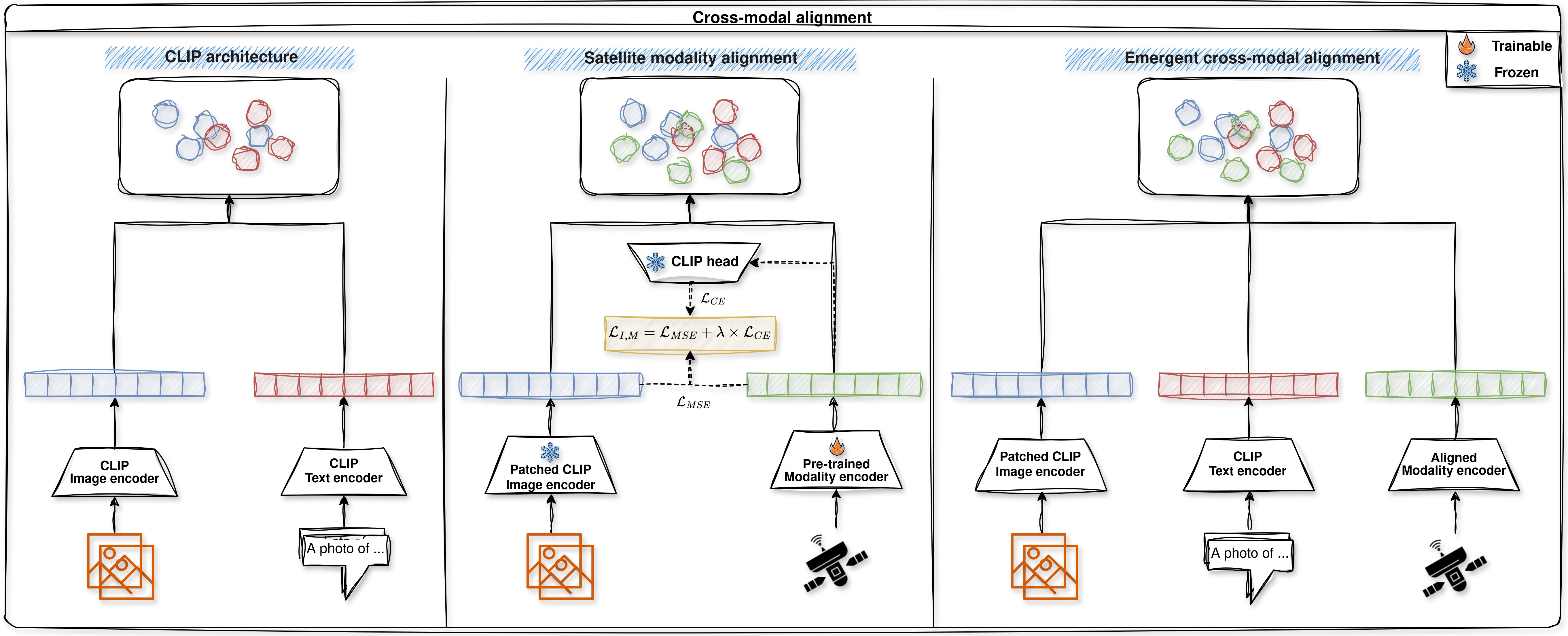}
    \caption{The second stage of our method concerns the cross-modal alignment of a satellite modality encoder within the shared embeddings space with CLIP visual and textual encoders. The image depicts: (A) Standard CLIP~\citep{radford2021learning} architecture with two separate image and text encoders projecting both embeddings into a shared embedding space. (B) Satellite modality alignment, introducing a pre-trained satellite modality encoder alongside the patched CLIP image encoder. The alignment is achieved through a joint loss function $L_{I,M} = L_{MSE} + \lambda \times L_{CE}$, where $L_{MSE}$ is the Mean Squared Error loss between the satellite modality encoder and the patched CLIP visual encoder embeddings, $L_{CE}$ is the Cross-Entropy loss between the satellite modality encoder embeddings and the output of CLIP’s text encoder as a frozen classification head to map satellite imagery features to the space of classes and $\lambda$ is a scaling factor balancing the two loss terms. (C) Emergent cross-modal alignment, showing the alignment between the patched CLIP image encoder, CLIP text encoder, and the newly aligned satellite modality encoder. All in all, this visualization illustrates the progression from the original CLIP architecture to a multi-modal shared embeddings space model which incorporates satellite imagery and maintains the initial CLIP shared embeddings space properties without the reliance on textual descriptions, without introducing any task-specific parameters, without training from scratch and without catastrophic forgetting.}
    \label{fig:align_clip}
\end{figure*}

In the cross-modal alignment stage, shown in Fig.~\ref{fig:align_clip}, we leverage the patched CLIP image encoder $\mathcal{M}_{\text{patched}}$, used in the first stage, as the teacher network, and a pre-trained satellite modality encoder $\mathcal{M}_{\text{sat}}$ as the student network. To effectively align these two networks, we adopt a straightforward approach grounded in the assumption that diverse modalities associated with the same sample should yield similar embeddings within the shared CLIP embedding space. The process involves a pair of modalities $\mathcal{I}_{RGB}$ and $\mathcal{I}_{RS}$ corresponding to RGB composites and some other satellite modality. For a given image $x_i \in \mathcal{I}_{RGB}$ and its corresponding sample $\widetilde{x}_i \in \mathcal{I}_{RS}$ from the two modalities, we obtain their respective embeddings $E_i = \mathcal{M}_{\text{patched}}(x_i)$ and $\widetilde{E}_i = \mathcal{M}_{\text{sat}}(\widetilde{x}_i)$. In cases where the embedding dimensions of $E_i$ and $\widetilde{E}_{i}$ differ, we introduce a linear projection head for the student network $\mathcal{M}_{\text{sat}}$ to ensure matching output embedding sizes.

Inspired by the knowledge distillation~\citep{hinton2015distilling} loss function, as well as previous distillation works~\citep{fang2021compressing, yang2020knowledge, kim2021comparing, yang2023clipkd}, the student is guided to mimic the teacher’s visual and textual embeddings via a joint objective function formulated as the linear combination of cross-entropy $\mathcal{L}_{CE}$ and mean squared error loss $\mathcal{L}_{MSE}$. 
\begin{equation} 
    \label{eq:alignment_loss}
    \mathcal{L}_{I,M} = \mathcal{L}_{MSE}(E_i, \widetilde{E}_{i}) + \lambda \times \mathcal{L}_{CE}(y_{pred},y_{i}).
\end{equation}
$\mathcal{L}_{CE}$ is determined by labeled data ($y_{i}$) supervision using the frozen output of CLIP's text encoder as an anchor to map image features to the space of classes, while $\mathcal{L}_{MSE}$ encourages the student to imitate the output embeddings of the teacher, in contrast to the softened class scores in the case of the original knowledge distillation loss. This alignment process effectively transfers the rich semantic understanding and cross-modal capabilities of CLIP to the domain of remote sensing.

\section{Experiments}

\subsection{Datasets} \label{sec:datasets}
\begin{figure}[tbh!]
    \centering
    \includegraphics[width=1.0\linewidth]{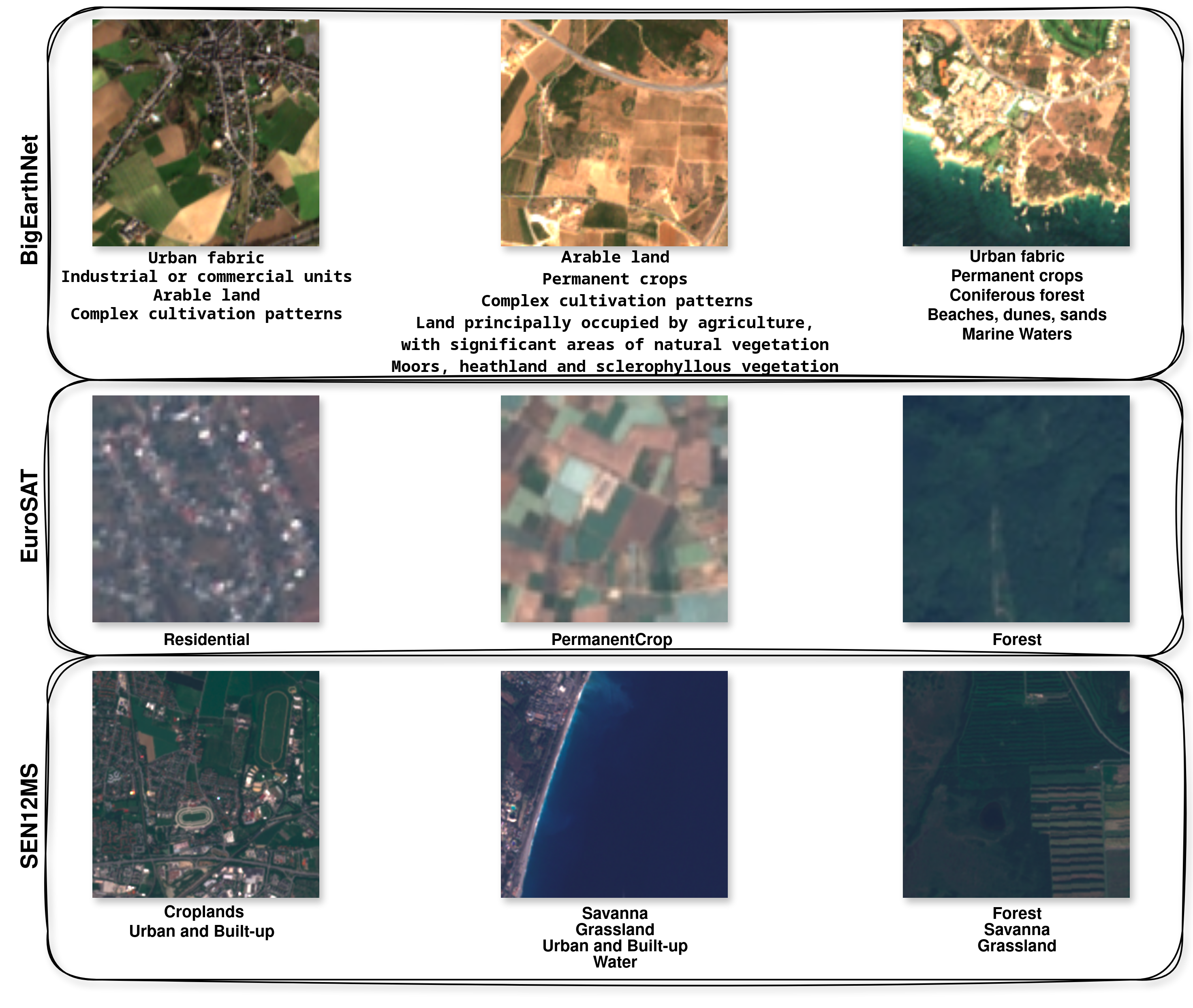}
    \caption{Comparison of image-label pairs sampled from the satellite datasets employed during the course of our experiments. Several innate features of the datasets, summarized in Table~\ref{tab:datasets}, can be observed. Specifically, the effect of atmospheric correction, which mitigate atmospheric interference thus enhancing true surface reflectance and reducing color distortion, can be distinguished when comparing BigEarthNet with EuroSAT and SEN12MS datasets. Moreover the spatial resolution differences are evident, as well as the variance with regards to the level of detail incorporated in each nomenclature.}
    \label{fig:datasets}
\end{figure}

We utilize three benchmark datasets for RS image classification, featuring a total of 5 different nomenclatures. These datasets vary in terms of the input satellite data specifications, but more importantly, they have adopted different labeling nomenclatures to describe the semantic content of the satellite images (see Fig.\ref{fig:datasets}).

\textbf{BigEarthNet}~\citep{sumbul2019bigearthnet} comprises of 590326 atmospherically corrected Sentinel-2 (S2-L2A) patches acquired between June 2017 and May 2018 over 10 European countries. The dataset is designed for multi-label LULC scene classification, each image patch is a section of i) $120\times120$ pixels for 10m bands; ii) $60\times60$ pixels for 20m bands; and iii) $20\times20$ pixels for 60m bands and is annotated with multiple land-cover classes sourced from the CORINE Land Cover (CLC) database of the year 2018~\citep{CLC2018}. The annotations are based on the detailed CORINE Level-3 class nomenclature (BigEarthNet-43), supplemented by an alternative 19-class LULC nomenclature (BigEarthNet-19) introduced by the authors~\citep{sumbul2020bigearthnet}, specifically for use in the machine-learning domain. To adopt a simplified classification scheme (BigEarthNet-5), CORINE Level-1 class nomenclature could be also utilized based on the CLC annotations. The dataset has been subsequently expanded to include Synthetic Aperture Radar Sentinel-1 (S1) patches, resulting in Sentinel-1 and Sentinel-2 patch pairs~\citep{sumbul2021bigearthnet}. Consequently, the term BigEarthNet now encompasses both the BigEarthNet-S1 and BigEarthNet-S2 datasets. Recently, BigEarthNet v2.0~\citep{clasen2024reben} was released featuring improvements along the lines of (a) atmospheric correction tool updates, (b) LULC label noise elimination and (c) dataset splits spatial de-correlation, aiming to enhance the dataset's overall quality and reliability. All experiments in this study utilized the BigEarthNet v1.0 dataset, hereafter referred to as BigEarthNet.

\textbf{EuroSAT}~\citep{helber2019eurosat} consists of 27000 single-date non-atmospherically corrected Sentinel-2 (S2-L1C) patches, distributed all over Europe, of size $64\times64$ pixels. The dataset is designed for multi-class LULC classification tasks and features a simple 10-class nomenclature. All 27000 patches have been manually checked.

\textbf{SEN12MS}~\citep{Schmitt2019} comprises of 180662 patch triplets that include Sentinel-1 dual-polarization SAR data, non-atmospherically corrected Sentinel-2 (S2-L1C) multi-spectral images and MODIS-derived land cover maps. These patches are of size 256$\times$256 pixels and are geographically distributed across Earth's inhabited continents while encompassing all four meteorological seasons. The dataset is designed for multi-label, multi-class LULC scene classification and semantic segmentation tasks. The four-band MODIS land cover patches have been created using 2016 data, at an upsampled pixel spacing of 10m. The first band contains land cover following the International Geosphere-Biosphere Programme (IGBP) classification scheme~\citep{loveland1997international}, while the remaining bands contain the Land Cover Classification System (LCCS) land cover, land use and hydrology layers~\citep{di2005land}. The authors have ultimately redesigned their dataset~\citep{Schmitt2021}, featuring a simplified IGBP scheme~\citep{yokoya2020report} for single-label and multi-label scene classification tasks, in order to ensure comparability with other land cover schemes and mitigate, to some extent, the class imbalance of their dataset.

\begin{table*}[tbh!]
    \centering
    \setlength{\tabcolsep}{12pt}
    \renewcommand{\arraystretch}{1.4}
    \resizebox{\linewidth}{!}{%
    \begin{tabular}{|c||c|c|c|c|c|c|c|}
        \hline
        \textbf{Dataset} & \textbf{\# of samples} & \textbf{Patch size} & \textbf{Modalities} & \textbf{Spatial Coverage} & \textbf{Temporal Coverage} & \textbf{Atmospherically Corrected} & \textbf{Nomenclature} \\
        \hline\hline
        BigEarthNet & 590,326 & 120$\times$120 & S1, S2 & Europe & Single season per patch & True & CORINE LULC L1, L2, L3 \\
        \hline
        SEN12MS & 180.661 & 256$\times$256 & S1, S2 & Global & Single season per patch & False & MODIS IGBP, LCCS \\
        \hline
        EuroSAT & 27.000 & 64$\times$64 & RGB, S2 & Europe & Single season per patch & False & 10 generic LULC classes \\
        \hline
    \end{tabular}
    }
    \caption{Synopsis of the fundamental traits of the satellite datasets employed during the course of our experiments.}
    \label{tab:datasets}
\end{table*}

\subsection{Experimental Setup}
\label{ssec:experimental_setup}

\textbf{Tasks.} Following~\citep{ilharco2022patching}, tasks are categorized as patching and supported tasks. Out of the diverse set of image classification tasks from~\citep{radford2021learning}, we use ImageNet~\citep{deng2009imagenet} as a representative supported task and BigEarthNet-S2~\citep{sumbul2021bigearthnet} as a representative patching task for satellite scene classification.

\textbf{Prompts.}
For each downstream dataset, we use a set of pre-defined prompts for each class, which we collected from prior works. We primarily use the prompts assembled by ~\citep{radford2021learning} for each downstream dataset, so that we provide a fair comparison of our results. In the case of satellite imagery datasets that were not present in the CLIP collection of downstream datasets (e.g. BigEarthNet), we use the alternative set of prompts proposed by ~\citep{zhai2022lit} for satellite datasets. Our approach follows the methodology introduced by~\citep{radford2021learning}, commonly referred to as ”prompt ensembling” and includes the generation of multiple candidate captions for each class, followed by the averaging of their representations prior to computing similarities.

\textbf{RGB Composites.}
In our experimental setup, it's important to note that all datasets utilized, with the exception of EuroSAT, did not contain pre-generated RGB composites. Instead, they provided the raw Sentinel-2 spectral bands. As a consequence, we had to produce the RGB composites ourselves. For Sentinel-2 imagery, we generated these composites using bands B4 (red), B3 (green), and B2 (blue). Following the ESA Copernicus guidelines\footnote{https://sentinel.esa.int/web/sentinel/user-guides/sentinel-2-msi/definitions}, we performed saturation mapping of the original reflectance values from these bands into 0-255 RGB digital values. This process involved scaling the raw sensor data to a standardized range suitable for visual display and analysis. The resulting composite image provides a natural color representation of the study area, enhancing the visual interpretation of land cover features and facilitating further RS analyses. Similarly, for other RS modalities, one could employ additional composites such as false-color composites or even multiple composite types for the patching and alignment stages, depending on the specific sensor characteristics. In all cases, it's crucial to adhere to the mapping guidelines provided by the respective mission and adapt the text prompts accordingly to accurately reflect the composite information represented.
 
\textbf{Models.} During the course of our experiments, we employ CLIP~\citep{radford2021learning} pre-trained vision transformer (ViT) models~\citep{dosovitskiy2020image}. We deliberately disregarded the ResNet-based models, in favor of the ViT-based models, predicated on prior research indicating that ResNets are more susceptible to catastrophic forgetting~\citep{ramasesh2021effect, ilharco2022patching}, while such interpolated networks suffer from variance collapse~\citep{jordan2022repair}. For the satellite modality encoder we utilize the SSL4EO-S12~\citep{wang2022ssl4eo} ViT-S-16 model architecture and weights, pre-trained via self-supervision using MoCo-v3~\citep{chenempirical}.

\textbf{Fine-tuning on patching task.} We fine-tune CLIP, as shown in Fig.~\ref{fig:patching_clip}, using RGB composites from the BigEarthNet-S2 dataset, with a batch size of 128 for 1 epoch using learning rate 1e-5 with 200 warm-up steps with a cosine annealing learning rate schedule and the AdamW optimizer~\citep{kingma2014adam, loshchilov2017decoupled} with weight decay~\citep{andriushchenko2023need} of 0.5. Additionally, we randomly apply ColorJitter, RandomGrayScale and GaussianBlur as data augmentations, aiming to compensate for the color variability between true-color images of surface reflectance and top-of-the-atmosphere reflectance products. During fine-tuning, we use CLIP’s text tower output, as the frozen final classification layer, so that we do not introduce any additional learnable parameters.

\textbf{Cross-modal alignment.} For the cross-modal alignment stage, we use a teacher-student framework, to align a pre-trained satellite (Sentinel-2) encoder with the CLIP image and text encoders, where the satellite encoder is the trainable student network and CLIP is the frozen teacher network, as shown in Fig. \ref{fig:align_clip}. We use again CLIP’s text tower output, as a frozen final classification layer, so that we do not introduce any additional learnable parameters. We guide the student to mimic teacher features, using paired Sentinel-2 multi-spectral data and RGB composites from the BigEarthNet-S2 dataset, with a batch size of 256 for 5 epochs using learning rate 1e-4 with 2000 warm-up steps with a cosine annealing learning rate schedule and the AdamW optimizer with weight decay of 0.01. We use the loss function listed in Eq.~\ref{eq:alignment_loss}, where we empirically set $\lambda=0.05$ in order to compensate for the difference in terms of magnitude between the two loss terms, while also purposely prioritizing slightly the MSE term over CE for the first few epochs, as discussed in Section~\ref{sec:discussion}.

\textbf{Evaluation.} We follow the setup of ~\citep{radford2021learning}. For the assessment of text-based zero-shot tasks, Mean Average Precision (mAP) and Accuracy (acc) are employed as the primary evaluation metrics, for multi-label and multi-class datasets, respectively. For each dataset, a set of pre-defined prompts for each class is utilized, as discussed in Section~\ref{ssec:experimental_setup}. 
In order to compute the embedding of each class, the embeddings of the pre-defined prompts are first computed using the text encoder, then they are averaged and L2-normalized. Each image is classified based on the cosine similarity of the L2-normalized image embedding with the class embeddings. 
For the evaluation of cross-modal retrieval tasks, we utilize Recall@K (R@k) as our performance metric. For two modalities, $A$ and $B$, we compute $A \rightarrow B$ scores using the cosine similarity between modalities $A$ and $B$ and rank the top-k samples of modality $B$ for each sample of modality $A$.

\begin{figure}[h]
    \centering
    \includegraphics[width=0.9\linewidth]{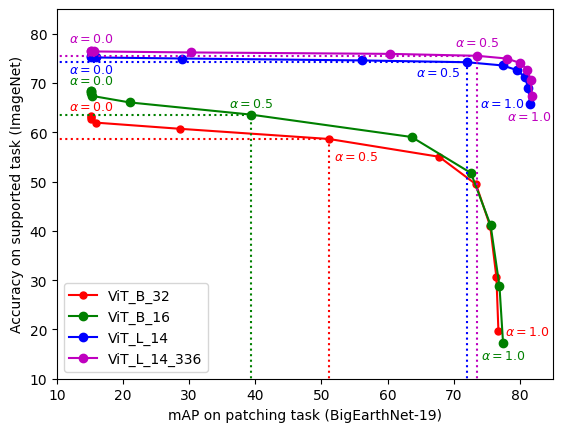}
    \caption{Performance trade-off between the patching task (BigEarthNet-19) and the representative supported task (Imagenet), while interpolating the fine-tuned and zero-shot weights for all four CLIP~\citep{radford2021learning} ViT variants. According to PAINT~\citep{ilharco2022patching}, we use a linear mixing coefficient $\alpha \in [0, 1]$ aiming to find an linear combination $W_{patch} = (1-\alpha) \times W_{zs} + \alpha \times W_{ft}$ of the fine-tuned $W_{ft}$ and zero-shot $W_{zs}$ weights where accuracy improves on the patching task (BigEarthNet-19) without reducing accuracy on the supported task (ImageNet). Essentially, for $\alpha=0.0$ we are using just the original zero-shot weights $W_{zs}$, for $\alpha=0.5$ we are averaging the zero-shot $W_{zs}$ and the fine-tuned $W_{ft}$ weights and for $\alpha=1.0$ we are using solely the fine-tuned weights $W_{ft}$. We have highlighted with dotted lines the performance for both supported (ImageNet) and patching (BigEarthNet-19) tasks for $\alpha=0.5$, which we utilized during the course of our patching experiments. Our results match with the authors observation that larger models are easier to patch since less movement is required to fit new data.}
    \label{fig:patching_eval}
\end{figure} 

\subsection{Results}

The first stage of Algorithm~\ref{alg:alignment} consists of patching CLIP by linearly interpolating weights (PAINT)~\citep{ilharco2022patching}. The goal is to improve accuracy on satellite scene classification tasks where the model performs poorly (patching tasks), without degrading performance on tasks where accuracy is already adequate (supported tasks).

Fig.~\ref{fig:patching_eval} summarizes the effect of patching using the BigEarthNet-19 dataset by highlighting the patching and supported tasks (ImageNet) performance trade-off for different mixing coefficients. We notice that the performance for both patching and supported tasks maximizes for $\alpha=0.5$. We report in detail the patching results for all four CLIP ViT-based models for the patching task (i.e. BigEarthNet-19) and the representative supported task (i.e. ImageNet) for the different $\alpha$ weight interpolation coefficients in \ref{ap:patchedmodelszeroshot}. For completeness, we also include the zero-shot performance evaluation results of the patched models, for the different $\alpha$ coefficients, on a set of diverse satellite scene classification nomenclatures and datasets. 

\begin{figure}[h]
    \centering
    \includegraphics[width=1.0\linewidth]{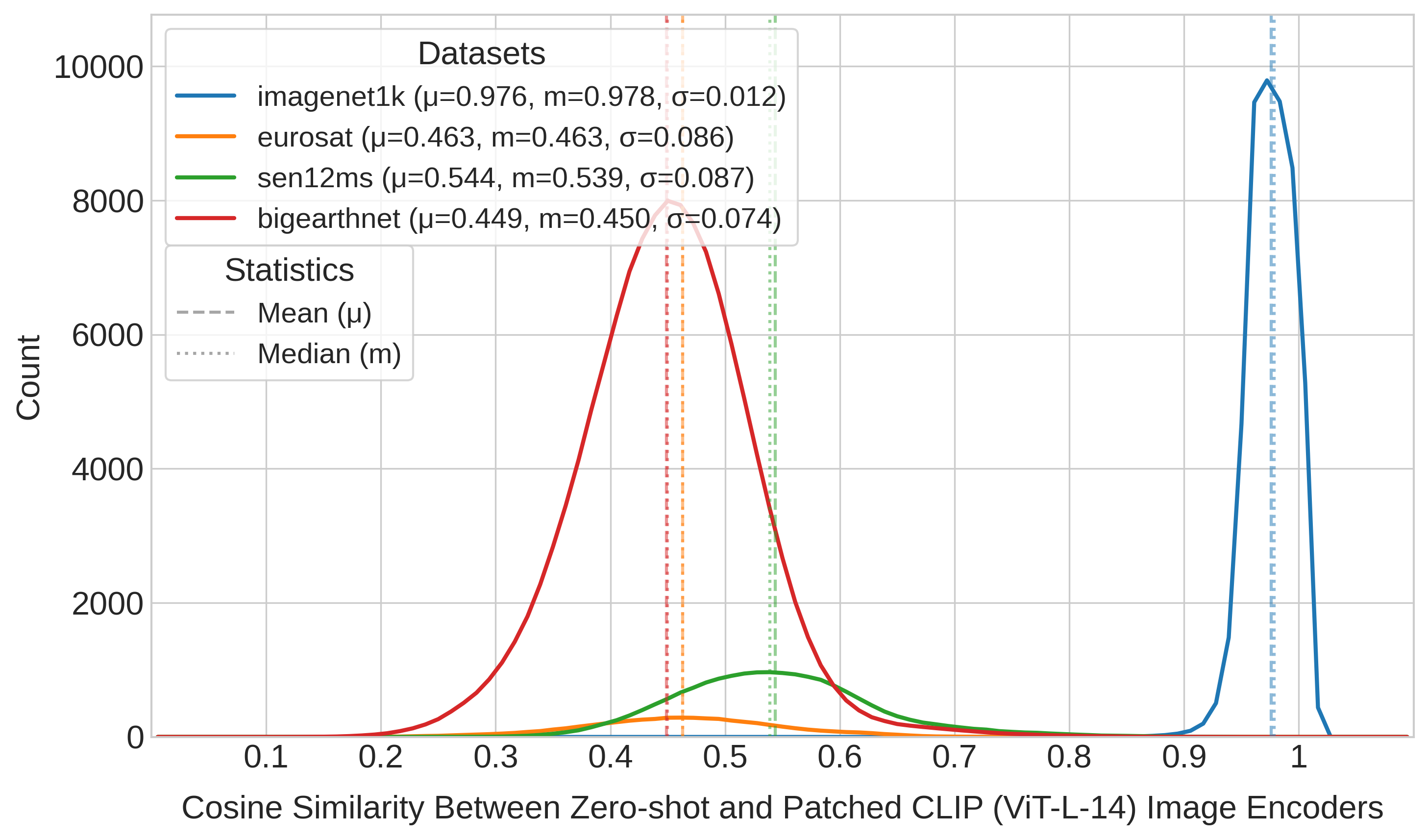}
    \caption{Distributions of cosine similarity between image embeddings from the original zero-shot CLIP (ViT-L-14) image encoder and the patched CLIP (ViT-L-14) image encoder. The distributions are shown for ImageNet1k, EuroSAT, SEN12MS, and BigEarthNet test sets. Dashed vertical lines represent the mean (µ) and dotted vertical lines represent the median (m) of each distribution. The high cosine similarity for ImageNet1k (µ=0.976, m=0.978) indicates minimal change to the embedding space for this supported task. In contrast, the significantly lower cosine similarities for the remote sensing datasets (EuroSAT: µ=0.463, m=0.463; SEN12MS: µ=0.544, m=0.539; BigEarthNet: µ=0.449, m=0.450) demonstrate a substantial shift in the embedding space, reflecting the adaptation of the patched model to the remote sensing domain.}
    \label{fig:encoder_similarity}
\end{figure}

\begin{figure*}
    \centering
    \includegraphics[width=1\linewidth]{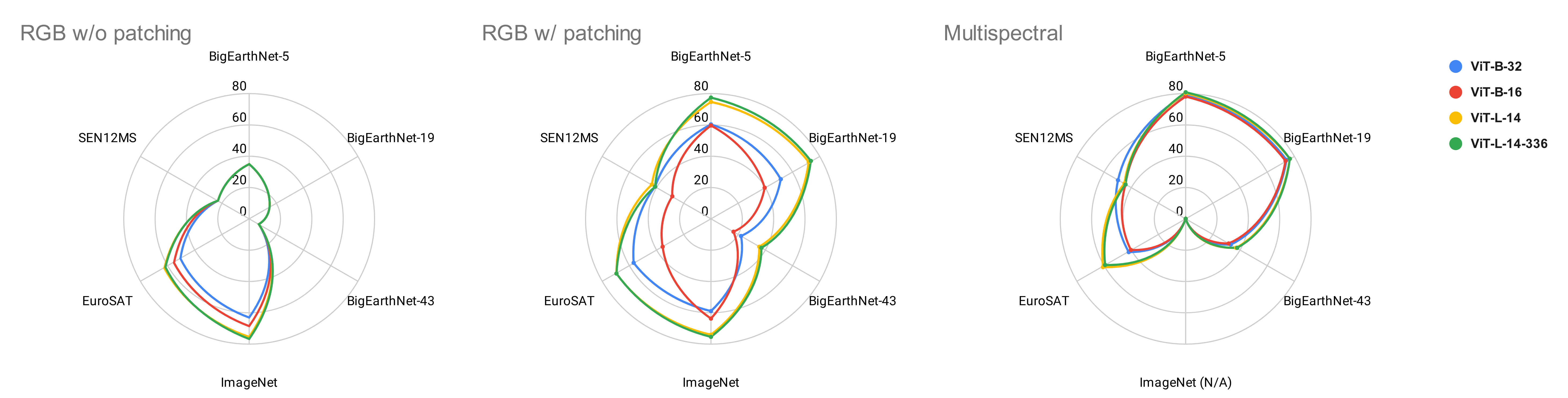}
    \caption{Comparison of zero-shot performance across different CLIP~\citep{radford2021learning} ViT variants on diverse satellite scene classification datasets and nomenclature. The radar charts illustrate performance in terms of either Accuracy (acc) or mean Average Precision (mAP) on six datasets: Imagenet (acc), BigEarthNet-5 (mAP), BigEarthNet-19 (mAP), BigEarthNet-43 (mAP), EuroSAT (acc), and SEN12MS (mAP). Left: Performance of CLIP ViT-based models without patching. Center: Performance after patching CLIP ViT-based encoders on BigEarthNet-19. Right: Performance of the ViT-S-16 satellite encoder after the cross-modal alignment with each one of the CLIP ViT variants, showing improved results on multispectral data. The charts demonstrate how patching and cross-modal alignment enhance CLIP's ability to classify satellite imagery across various datasets and nomenclatures.}
    \label{fig:patching_diff}
\end{figure*}

Fig.~\ref{fig:patching_diff} compares and contrasts the pre- and post-patching performance for the patching and the representative supported tasks and the various nomenclatures and datasets across all candidate models. Best performance trade-off across patching and supported tasks is achieved using the larger ViT-L-14 and ViT-L-14-336 models. These results, are in-line with those of \citet{ramasesh2021effect} and \citet{mehta2023empirical} who observed that larger models are less susceptible to catastrophic forgetting and \citet{ilharco2022patching} who highlighted that PAINT is more effective for larger models. Fig.~\ref{fig:encoder_similarity} provides a more detailed analysis of the patching process's impact by visualizing the distribution of cosine similarities between image embeddings from the original zero-shot CLIP (ViT-L-14) image encoder and the patched CLIP (ViT-L-14) image encoder. This analysis is performed on the test sets of all four datasets: ImageNet1k, EuroSAT, SEN12MS, and BigEarthNet. We focus solely on the image encoder because, as per the PAINT protocol~\citep{ilharco2022patching}, only the image encoder's weights are modified during patching; the text encoder retains its original, pre-trained weights. The cosine similarity between the original and patched embeddings remains extremely high for ImageNet1k (µ=0.976, m=0.978), indicating minimal disruption to the representation of this supported task. In contrast, the distributions for the remote sensing datasets shift significantly towards lower cosine similarity values. This substantial change in the embedding space for the RS datasets demonstrates the successful adaptation of the patched model to the remote sensing domain, explaining the performance improvements reported. To provide a comprehensive view, we compare and contrast in Table~\ref{tab:clip_variants_comparison} the classification performance of our best patched model, i.e. ViT-L-14, with related RS-tailored CLIP ViT-L-14 variants mentioned in the Section~\ref{sec:related_work}, i.e. RemoteCLIP~\citep{liu2023remoteclip}, GeoRSCLIP~\citep{zhang2023rs5m}, SkyCLIP-50 and CLIP-LAION-RS~\citep{wang2024skyscript}, as well as with the original CLIP~\citep{radford2021learning}, across diverse satellite scene classification datasets. To ensure a fair and unbiased comparison with the aforementioned models, we adhered to the exact text prompts specified in their respective original publications. Our Patched CLIP excels on almost every RS dataset, while maintaining competitive performance on ImageNet. Notably, it achieves the highest scores on 5 out of 6 datasets.

\begingroup
\setlength{\tabcolsep}{24pt}
\renewcommand{\arraystretch}{1.2}
\begin{table*}[tbh!]
    \centering
    \resizebox{\linewidth}{!}{%
    \begin{tabular}{|c||c|c|c|c|c||c|c|c|c|c|}
        \hline \multirow{2}{*}{\textbf{ViT-B-32}} & \multicolumn{5}{c||}{\textbf{RGB$\rightarrow$MS}} & \multicolumn{5}{c|}{\textbf{MS$\rightarrow$RGB}} \\
        \cline{2-11} & \textbf{R@1} & \textbf{R@5} & \textbf{R@10} & \textbf{R@20} & \textbf{R@50} & \textbf{R@1} & \textbf{R@5} & \textbf{R@10} & \textbf{R@20} & \textbf{R@50} \\
        \hline\hline BigEarthNet & 16.71 & 36.50 & 46.90 & 57.94 & 72.59 & 3.85 & 11.44 & 16.91 & 24.18 & 36.98 \\
        \hline EuroSAT & 1.67 & 5.28 & 8.44 & 13.00 & 22.20 & 0.37 & 1.81 & 3.89 & 6.61 & 13.65 \\
        \hline SEN12MS & 0.39 & 1.54 & 2.60 & 4.39 & 8.42 & 0.19 & 0.96 & 1.81 & 3.28 & 6.45 \\
        \hline\hline \multirow{2}{*}{\textbf{ViT-B-16}} & \multicolumn{5}{c||}{\textbf{RGB$\rightarrow$MS}} & \multicolumn{5}{c|}{\textbf{MS$\rightarrow$RGB}} \\
        \cline{2-11} & \textbf{R@1} & \textbf{R@5} & \textbf{R@10} & \textbf{R@20} & \textbf{R@50} & \textbf{R@1} & \textbf{R@5} & \textbf{R@10} & \textbf{R@20} & \textbf{R@50} \\
        \hline\hline BigEarthNet & 12.68 & 30.42 & 40.57 & 51.95 & 67.51 & 2.88 & 8.82 & 13.33 & 19.71 & 31.24 \\
        \hline EuroSAT & 1.89 & 6.37 & 10.04 & 15.22 & 26.63 & 1.00 & 3.59 & 6.33 & 10.39 & 18.83 \\
        \hline SEN12MS & 0.48 & 1.60 & 2.66 & 4.56 & 9.05 & 0.25 & 1.03 & 1.96 & 3.33 & 6.64 \\
        \hline\hline \multirow{2}{*}{\textbf{ViT-L-14}} & \multicolumn{5}{c||}{\textbf{RGB$\rightarrow$MS}} & \multicolumn{5}{c|}{\textbf{MS$\rightarrow$RGB}} \\
        \cline{2-11} & \textbf{R@1} & \textbf{R@5} & \textbf{R@10} & \textbf{R@20} & \textbf{R@50} & \textbf{R@1} & \textbf{R@5} & \textbf{R@10} & \textbf{R@20} & \textbf{R@50} \\
        \hline\hline BigEarthNet & 24.86 & 47.43 & 57.37 & 66.88 & 78.49 & 6.14 & 16.42 & 23.30 & 31.87 & 45.40 \\
        \hline EuroSAT & 3.69 & 11.02 & 16.54 & 24.50 & 37.56 & 1.13 & 4.31 & 7.20 & 12.24 & 22.31 \\
        \hline SEN12MS & 0.82 & 3.06 & 5.01 & 8.48 & 15.55 & 0.27 & 1.12 & 2.10 & 3.69 & 8.04 \\
        \hline\hline \multirow{2}{*}{\textbf{ViT-L-14-336}} & \multicolumn{5}{c||}{\textbf{RGB$\rightarrow$MS}} & \multicolumn{5}{c|}{\textbf{MS$\rightarrow$RGB}} \\
        \cline{2-11} & \textbf{R@1} & \textbf{R@5} & \textbf{R@10} & \textbf{R@20} & \textbf{R@50} & \textbf{R@1} & \textbf{R@5} & \textbf{R@10} & \textbf{R@20} & \textbf{R@50} \\
        \hline\hline BigEarthNet & 26.94 & 50.44 & 60.40 & 69.83 & 81.03 & 7.28 & 18.79 & 26.13 & 35.01 & 48.75 \\
        \hline EuroSAT & 3.76 & 10.87 & 16.70 & 24.65 & 38.13 & 1.13 & 3.96 & 6.56 & 10.15 & 19.41 \\
        \hline SEN12MS & 0.96 & 3.84 & 6.47 & 10.29 & 18.78 & 0.22 & 0.93 & 1.66 & 3.04 & 6.90 \\
        \hline
    \end{tabular}
    }
    \caption{Cross-modal retrieval performance of all four CLIP ViT-based models on various RS datasets. Results show RGB$\rightarrow$MS and MS$\rightarrow$RGB retrieval for BigEarthNet, EuroSAT, and SEN12MS datasets, with Recall@K metrics (K=1,5,10,20,50). Larger models generally outperform smaller ones, with BigEarthNet consistently yielding the highest scores. Importantly, MS$\rightarrow$RGB retrieval consistently outperforms RGB$\rightarrow$MS across all models and datasets, indicating that the multi-spectral modality provides more informative features for cross-modal retrieval tasks.}
    \label{tab:clip_retrieval}
\end{table*}
\endgroup

During the cross-modal alignment stage, we are distilling each one of the four CLIP ViT-based models into a pre-trained ViT-S-16 satellite (Sentinel-2) modality encoder. We summarize our results for zero-shot classification in Fig.~\ref{fig:patching_diff}. \ref{ap:bestmodel} contains more performance results per dataset and per class. When compared to both the non-patched and patched RGB CLIP variants, our results demonstrate notable performance improvements across all multi-spectral datasets used for evaluation, with the exception of the EuroSAT dataset. We argue that this is due to the objective function used during the alignment stage and further justify this claim in Section~\ref{sec:discussion}. For completeness, the cross-modal retrieval results are reported in Table~\ref{tab:clip_retrieval}, setting the baselines for the RGB$\rightarrow$MS and MS$\rightarrow$RGB retrieval tasks evaluation. It is evident that significantly better performance is achieved when aiming for a more informative modality during the cross-modal retrieval, i.e., RGB imagery aiming to retrieve the respective multi-spectral imagery. To thoroughly evaluate the quality of representations learned by our aligned models, we conducted linear probing experiments involving various RS classification datasets. All models are trained for 20 epochs. Table~\ref{tab:foundation_models_comparison} presents these results, comparing our aligned model to several state-of-the-art pre-trained foundation models. Our aligned modality encoder excels on almost every RS dataset, proving the quality of representation learned during the cross-modal alignment stage of our method.

\begingroup
\setlength{\tabcolsep}{24pt}
\renewcommand{\arraystretch}{1.2}
\begin{table*}[tbh!]
    \centering
    \resizebox{\linewidth}{!}{%
    \begin{tabular}{|c||c|c|c|c|c|c|}
        \hline \textbf{Classification} & \textbf{CLIP} & \textbf{RemoteCLIP} & \textbf{GeoRSCLIP} & \textbf{SkyCLIP-50} & \textbf{CLIP-LAION-RS} & \textbf{Patched CLIP (ours)} \\
        \hline\hline BigEarthNet-5 & 35,16 & 35,26 & 36,77 & 35,16 & 35,16 & \textbf{74,92} \\
        \hline BigEarthNet-19 & 15,18 & 15,26 & 17,44 & 15,18 & 15,18 & \textbf{72,04} \\
        \hline BigEarthNet-43 & 6,83 & 6,85 & 8,39 & 6,82 & 6,82 & \textbf{35,26} \\
        \hline ImageNet & 75,49 & 67,95 & 65,97 & 75,54 & \textbf{75,57} & 74,22 \\
        \hline EuroSAT & 63,72 & 59,94 & \textbf{74,83} & 70,78 & 71,22 & 69,70 \\
        \hline SEN12MS & 22,83 & 22,83 & 24,06 & 22,83 & 22,83 & \textbf{43,44} \\
        \hline
    \end{tabular}
    }
    \caption{Classification performance comparison of RS-tailored CLIP ViT-L-14 variants across diverse satellite scene classification datasets, with the highest score for each dataset highlighted in bold. The table presents percentage scores (\%), for RGB setting, in terms of either Accuracy (acc) or mean Average Precision (mAP) on six datasets: Imagenet (acc), BigEarthNet-5 (mAP), BigEarthNet-19 (mAP), BigEarthNet-43 (mAP), EuroSAT (acc), and SEN12MS (mAP). We evaluate various CLIP models, including the original CLIP~\citep{radford2021learning} and RS-specialized variants, i.e. RemoteCLIP~\citep{liu2023remoteclip}, GeoRSCLIP~\citep{zhang2023rs5m}, SkyCLIP-50 and CLIP-LAION-RS~\citep{wang2024skyscript} and our proposed Patched CLIP. Our Patched CLIP excels on almost every RS dataset, while maintaining competitive performance on ImageNet.}
    \label{tab:clip_variants_comparison}
\end{table*}
\endgroup

\begingroup
\setlength{\tabcolsep}{8pt}
\renewcommand{\arraystretch}{1.2}
\begin{table*}[tbh!]
    \centering
    \resizebox{\linewidth}{!}{%
    \begin{tabular}{|c||c|c|c|c|c|c|c|c|}
        \hline \textbf{Classification} & \textbf{Random Init. (ViT-B)} & \textbf{SSL4EO-MoCo (ViT-S)} & \textbf{SatMAE (ViT-B)} & \textbf{CROMA (ViT-B)} & \textbf{DOFA (ViT-B)} & \textbf{DOFA (ViT-L)} & \textbf{Ours (ViT-S)} \\
        \hline\hline
        BigEarthNet-5 & 65.78 & 83.36 & 80.22 & 84.34 & 78.02 & 80.47 & \textbf{90.39} \\
        \hline
        BigEarthNet-19 & 50.97 & 65.91 & 61.11 & 66.43 & 61.85 & 62.76 & \textbf{77.50} \\
        \hline
        BigEarthNet-43 & 28.03 & 42.07 & 39.56 & 42.92 & 40.16 & 40.89 & \textbf{52.76} \\
        \hline
        EuroSAT & 84.33 & 91.33 & 84.41 & 89.13 & 92.01 & \textbf{92.78} & 91.36 \\
        \hline
        SEN12MS & 39.91 & 71.93 & 54.88 & 59.03 & 60.77 & 63.07 & \textbf{73.28} \\
        \hline
    \end{tabular}
    }
    \caption{Comparison of linear probing results for different pre-trained foundation models on various RS classification tasks, for multi-spectral setting. All models are trained for 20 epochs. Reported metrics include Accuracy (acc) for EuroSAT and mean Average Precision (mAP) for the other datasets, all presented as percentages (\%). Backbones compared include Random Init., SSL4EO-MoCo~\citep{wang2022ssl4eo}, SatMAE~\citep{cong2022satmae}, CROMA~\citep{fuller2024croma}, DOFA~\citep{xiong2024neural}, and the proposed method (Ours), utilizing ViT-B, ViT-S, and ViT-L backbones as indicated. Bold values highlight the best performance for each dataset. Our aligned modality encoder excels on almost every RS dataset, proving the quality of representation learned during the cross-modal alignment stage of our method.}
    \label{tab:foundation_models_comparison}
\end{table*}
\endgroup

\subsection{Ablation Study}
In this section we systematically investigate various design choices for each one of the two stages of the proposed method. Unless explicitly stated, the experimental setup is similar to Section~\ref{ssec:experimental_setup}. We conduct the entirety of the ablation study using the smallest pre-trained CLIP model variant, namely ViT-B-32.

\begin{figure}[tbh!]
    \centering
    \includegraphics[width=1\linewidth]{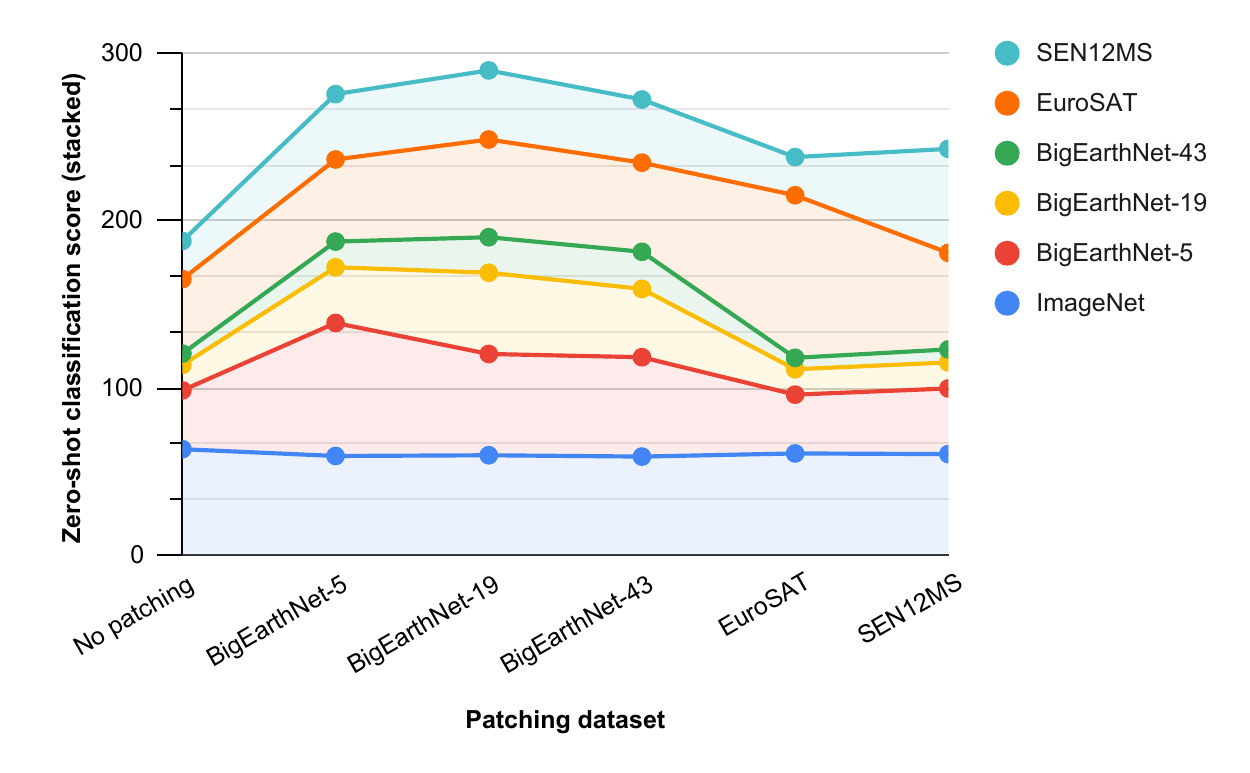}
    \caption{This graph illustrates the results of our ablation study evaluating how the choice of patching dataset affects CLIP's zero-shot classification performance. The horizontal axis shows the different patching datasets used, including a 'No patching' baseline. The vertical axis presents the stacked cumulative classification scores (mean Average Precision or Accuracy depending on the dataset) for each patching dataset. Each line corresponds to a different evaluation dataset described comprehensively in Section~\ref{sec:datasets}, showing how performance on these datasets changes based on the patching dataset used. The graph reveals that patching with BigEarthNet-19 yields the best overall improvement across the different datasets considered.}
    \label{fig:patching_stacked}
\end{figure}

\textbf{Patching dataset selection.} 
The dataset used for patching CLIP is arguably the single most important factor in the first stage of Algorithm~\ref{alg:alignment}. To this end, we evaluate the effect of patching CLIP and the resulting zero-shot performance on both supported and patching tasks, using each one of the datasets considered in Section~\ref{sec:datasets} as the proxy task. 
We present the results of this ablation study stacked for each patching dataset in Fig.~\ref{fig:patching_stacked}. Patching with the EuroSAT dataset shows the weakest performance, offering improvements only to itself and none to the other datasets. Patching with SEN12MS results in minor improvements for other datasets and a significant increase, as expected, for itself. Patching with any variation of the BigEarthNet dataset (5/19/43-class nomenclature) as the proxy task, outperforms the EuroSAT and the SEN12MS datasets in terms of both supported and patching tasks evaluation. Arguably the significantly bigger dataset size of BigEarthNet compared to SEN12MS and EuroSAT plays a pivotal role to the outcome of the patching procedure. Moreover there is a clear correlation between the number of classes in the featured datasets and the model's improved generalization. This relationship is expected, as a greater variety of classes allows the model to learn more diverse and abstract visual representations, grounded in the corresponding textual data. However, for a more accurate interpretation of our results when patching using BigEarthNet-19 and BigEarthNet-43, it's crucial to consider the severe class imbalances that emerge when utilizing the BigEarthNet 43-class nomenclature. Interestingly, our results can not be considered conclusive regarding the impact of multi-label or multi-class dataset nature on patching outcomes. This inconclusiveness is primarily due to the coexisting differences in dataset sizes. Our initial expectation was that multi-label datasets would yield superior results, as each image in such datasets combines several abstract representations, more closely mirroring the complexity of the physical world. However, our current findings neither confirm nor disprove this hypothesis. Lastly, the native resolution of the dataset patches does not seem to significantly affect the patching outcome.

\textbf{Patching instead of just fine-tuning.}
Table~\ref{tab:finetuning_vs_patching} presents a quantitative analysis of the trade-off between adapting CLIP to the remote sensing domain and preserving its general zero-shot capabilities. We use BigEarthNet-19 as the adaptation dataset. The results clearly demonstrate that while standard fine-tuning diminishes performance on the representative supported task (ImageNet) across all CLIP model variants, the PAINT patching method effectively mitigates this catastrophic forgetting. Specifically, patching maintains ImageNet accuracy close to the original CLIP baseline, while substantially improving performance on the patching task. Furthermore, patching exhibits superior cross-dataset generalization compared to fine-tuning. While fine-tuning may lead to rapid convergence on the target task, it comes at the cost of catastrophic forgetting and overfitting to the specific task's narrow data distribution, undermining the foundational knowledge and broad applicability established through large-scale pre-training. This evidence supports the conclusion that PAINT preserves CLIP's robustness, acquired during large-scale pre-training, while simultaneously enabling the acquisition of domain-specific knowledge relevant to remote sensing imagery.

\begingroup
\setlength{\tabcolsep}{12pt}
\renewcommand{\arraystretch}{1.2}
\begin{table*}[t]
    \centering
    \resizebox{\linewidth}{!}{%
    \begin{tabular}{|c||c|c||c|c||c|c||c|c|}
        \hline \multirow{2}{*}{\textbf{Dataset}} & \multicolumn{2}{c||}{\textbf{ViT-B-32}} & \multicolumn{2}{c||}{\textbf{ViT-B-16}} & \multicolumn{2}{c||}{\textbf{ViT-L-14}} & \multicolumn{2}{c|}{\textbf{ViT-L-14-336}} \\
        \cline{2-9} & \textbf{Finetuned} & \textbf{Patched} & \textbf{Finetuned} & \textbf{Patched} & \textbf{Finetuned} & \textbf{Patched} & \textbf{Finetuned} & \textbf{Patched} \\
        \hline\hline BigEarthNet-19 & 76.76 & 51.13 & 77.46 & 39.36 & 81.53 & 72.04 & 81.86 & 73.53 \\
        \hline ImageNet & 19.74 & 58.65 & 17.32 & 63.59 & 65.72 & 74.22 & 67.29 & 75.53 \\
        \hline EuroSAT & 43.87 & 56.57 & 31.69 & 35.50 & 59.70 & 69.70 & 57.98 & 69.24 \\
        \hline SEN12MS & 46.80 & 40.93 & 43.27 & 28.65 & 47.58 & 43.44 & 47.58 & 40.78 \\
        \hline BigEarthNet-5 & 72.80 & 60.30 & 75.00 & 59.92 & 75.83 & 74.92 & 78.02 & 77.28 \\
        \hline BigEarthNet-43 & 33.01 & 21.97 & 32.48 & 16.75 & 39.08 & 35.26 & 40.05 & 37.03 \\ 
        \hline
    \end{tabular}
    }
    \caption{Zero-shot performance comparison of CLIP ViT-based models (ViT-B-32, ViT-B-16, ViT-L-14, and ViT-L-14-336) on remote sensing datasets (EuroSAT, SEN12MS, BigEarthNet-5, BigEarthNet-19, BigEarthNet-43) and ImageNet. The table contrasts the performance of models fine-tuned on BigEarthNet-19 with models patched using the BigEarthNet-19 dataset. Performance is evaluated using mean Average Precision (mAP) for multi-label datasets (BigEarthNet variants and SEN12MS) and accuracy (acc) for single-label datasets (ImageNet and EuroSAT). Results demonstrate that patching, contrary to fine-tuning, consistently improves performance on the target remote sensing datasets, while maintaining performance on the representative supported task (ImageNet).}
    \label{tab:finetuning_vs_patching}
\end{table*}
\endgroup

\textbf{Alignment loss variants.} 
The alignment across all modalities can be measured using cross-modal retrieval and text-based zero-shot tasks. To that end, we evaluate the effectiveness of various alignment loss variants on the cross-modal alignment. To cover every aspect, we also evaluate the ability of the patched CLIP model, as the teacher network during the satellite modality encoder alignment stage. We summarize our results in Table~\ref{tab:ablation_loss_classification} and Table~\ref{tab:ablation_loss_retrieval} for classification and retrieval tasks respectively. 

\begingroup
\setlength{\tabcolsep}{24pt}
\renewcommand{\arraystretch}{1.2}
\begin{table*}[h]
    \centering
    \resizebox{\linewidth}{!}{%
    \begin{tabular}{|c||c|c|c||c|c|c|}
    \hline \multirow{2}{*}{ \textbf{Classification} } & \multicolumn{3}{c||}{ \textbf{Non patched} } & \multicolumn{3}{c|}{ \textbf{Patched} } \\
    \cline { 2 - 7 } & \textbf{CE} & \textbf{MSE} & \textbf{MSE \& CE} & \textbf{CE} & \textbf{MSE} & \textbf{MSE \& CE} \\
    \hline\hline BigEarthNet-5 & 67.61 & 35.16 & 69.50 & 67.61 & 63.76 & 79.18 \\
    \hline BigEarthNet-19 & 81.68 & 15.18 & 67.81 & 81.68 & 54.11 & 74.69 \\
    \hline BigEarthNet-43 & 32.54 & 6.82 & 28.96 & 32.54 & 23.33 & 32.87 \\
    \hline EuroSAT & 35.59 & 41.91 & 40.76 & 35.59 & 39.89 & 42.11 \\
    \hline SEN12MS & 47.48 & 22.83 & 47.16 & 47.48 & 32.02 & 49.98 \\
    \hline
    \end{tabular}
    }
    \caption{Ablation study results, concerning the effectiveness of various alignment loss variants on the classification task. The left side of the table shows performance without patching (with the BigEarthNet-19 dataset) while the right side includes patching. The table presents results for three loss functions: Cross-Entropy (CE), Mean Squared Error (MSE), and a combination of MSE \& CE. Results demonstrate that patching generally improves performance, particularly when using the combined MSE \& CE loss. CE loss consistently outperforms MSE for the classification tasks in both patched and non-patched scenarios. These results should be interpreted in conjunction with the classification performance results presented in Table~\ref{tab:ablation_loss_retrieval} to gain a comprehensive understanding of the ablation study results.}
    \label{tab:ablation_loss_classification}
\end{table*}
\endgroup

\begingroup
\setlength{\tabcolsep}{4pt}
\renewcommand{\arraystretch}{1.2}
\begin{table*}[h]
    \centering
    \resizebox{\textwidth}{!}{
    \begin{tabular}{|c||c|c|c|c|c|c||c|c|c|c|c|c|}
    \hline \multirow{3}{*}{ \textbf{Retrieval} } & \multicolumn{6}{c||}{ \textbf{Non patched} } & \multicolumn{6}{c|}{ \textbf{Patched} } \\
    \cline{2-13} & \multicolumn{2}{c|}{ \textbf{CE} } & \multicolumn{2}{c|}{ \textbf{MSE} } & \multicolumn{2}{c||}{ \textbf{MSE \& CE} } & \multicolumn{2}{c|}{ \textbf{CE} } & \multicolumn{2}{c|}{ \textbf{MSE} } & \multicolumn{2}{c|}{ \textbf{MSE \& CE} } \\
    \cline{2-13} & \textbf{RGB$\rightarrow$MS} & \textbf{MS$\rightarrow$RGB} & \textbf{RGB$\rightarrow$MS} & \textbf{MS$\rightarrow$RGB} & \textbf{RGB$\rightarrow$MS} & \textbf{MS$\rightarrow$RGB} & \textbf{RGB$\rightarrow$MS} & \textbf{MS$\rightarrow$RGB} & \textbf{RGB$\rightarrow$MS} & \textbf{MS$\rightarrow$RGB} & \textbf{RGB$\rightarrow$MS} & \textbf{MS$\rightarrow$RGB} \\
    \hline\hline BigEarthNet & 0.12 & 0.24 & 40.01 & 12.05 & 27.91 & 16.91 & 0.65 & 2.45 & 65.07 & 27.88 & 46.90 & 16.91 \\
    \hline EuroSAT & 0.31 & 0.30 & 7.33 & 3.56 & 5.02 & 2.83 & 0.43 & 0.41 & 15.28 & 5.98 & 8.44 & 3.89 \\
    \hline SEN12MS & 0.10 & 0.08 & 2.77 & 3.08 & 2.77 & 1.48 & 0.16 & 0.13 & 4.85 & 3.08 & 2.60 & 1.81 \\
    \hline
    \end{tabular}
    }
    \caption{Ablation study results, concerning the effectiveness of various alignment loss variants on the cross-modal retrieval task. Cross-modal retrieval performance (R@10) of CLIP ViT-based models on satellite imagery datasets, comparing patched and non-patched approaches. Results are shown for RGB$\rightarrow$MS and MS$\rightarrow$RGB retrieval on BigEarthNet, EuroSAT, and SEN12MS datasets using three loss functions: Cross-Entropy (CE), Mean Squared Error (MSE), and combined MSE \& CE. The results demonstrate that patching generally improves retrieval performance, with MSE loss showing particular effectiveness for retrieval tasks. These results should be interpreted in conjunction with the classification performance results presented in Table~\ref{tab:ablation_loss_classification} to gain a comprehensive understanding of the ablation study results.}
    \label{tab:ablation_loss_retrieval}
\end{table*}
\endgroup

The results demonstrate both the necessity and the effectiveness of the initial patching stage. More specifically, we notice significant performance gains on the derived cross-modal alignment when using the patched CLIP model, for the alignment of the satellite modality encoder, instead of the original CLIP model weights. This can be observed both in the classification task (see Table~\ref{tab:ablation_loss_classification}) and the 
retrieval task (see Table~\ref{tab:ablation_loss_retrieval}).
In terms of the loss variants evaluated, we yield the best results for both classification and retrieval tasks, using the proposed loss of Eq.~\ref{eq:alignment_loss}. The exclusive utilization of either cross entropy (CE) or mean squared error (MSE) loss, for the satellite modality encoder alignment, results in a tendency for alignment with either the CLIP text or image encoder. More specifically, the aligned satellite modality encoder gravitates towards the CLIP text encoder when using the CE loss and towards the CLIP image encoder when using the MSE loss. This behavior disrupts the alignment between the satellite modality encoder and the residual modality.

\textbf{Assessing the utilization of multi-spectral information after cross-modal alignment.}
We assess whether the Sentinel-2 modality encoder, after the cross-modal alignment, learns to utilize the full multi-spectral information or merely reconstructs the RGB composite values from the corresponding bands. For this ablation study, we used the same pre-trained weights for the modality encoder, for both use-cases. In the RGB bands ablation, we loaded these pre-trained weights into an encoder architecture modified to accept only three input channels, representing the red, green, and blue spectral bands. Crucially, only the weights corresponding to the three RGB input channels of the pre-trained encoder were used; weights associated with other spectral bands were discarded. For the Sentinel-2, the pre-trained weights were loaded into an encoder with 13 input channels, corresponding to the full range of Sentinel-2 spectral bands. During alignment, we observed that the alignment loss for the RGB-only variant plateaued after just 2 epochs, indicating a failure to converge further. In contrast, the alignment loss for the all-bands variant continued to decrease, leading to the observed performance improvements. As shown in Table~\ref{tab:rgb_vs_ms} aligning with all Sentinel-2 bands significantly outperforms RGB-only alignment, confirming that our procedure successfully integrates complementary information from all multi-spectral bands, leverage information beyond RGB.

\textbf{Patching and Alignment of additional modalities.}
While our proposed method has demonstrated promising results for Sentinel-2 (S2) satellite imagery it is valuable to showcase its applicability and effectiveness to additional modalities, such as Sentinel-1 (S1). Besides, in a scaled-up version of our work, we would incorporate more modalities, as discussed in Section~\ref{sec:discussion}. To assess the performance of our method on Sentinel-1 data, we conducted this ablation study investigating three scenarios: (a) aligning Sentinel-1 without patching, (b) aligning S1 post-patching with S2, and (c) aligning S1 post-patching with S1. 

In our experiments, we maintain the same experimental setup mentioned in Section~\ref{ssec:experimental_setup} with the only differences being the utilization of a pre-trained S1 modality encoder instead of a pre-trained S2 modality encoder and the generation of false-color RGB composites from S1 data, according to Copernicus guidelines, instead of true-color RGB composites from S2 data. We deliberately disregarded the EuroSAT dataset from our results since it does not include Sentinel-1 imagery. To account for the false-color composites, we adapted our text prompts by incorporating the false-color property within the prompts, ensuring accurate interpretation of the input data. The results of our ablation study (shown in Table~\ref{tab:ablation_s1}) demonstrate that aligning S1 post-patching with S1 consistently outperforms the other two approaches. Furthermore, aligning S1 post-patching with S2 data yields better performance compared to aligning S1 without patching. These findings prove once more that the patching process is crucial for learning more robust RS representations compared to the original CLIP model, even when patching using a different modality.

\begingroup
\setlength{\tabcolsep}{24pt}
\renewcommand{\arraystretch}{1.2}
\begin{table*}[h]
    \centering
    \resizebox{\linewidth}{!}{%
    \begin{tabular}{|c||c|c|c|}
        \hline \textbf{Classification} & \textbf{Align S1 without Patching} & \textbf{Align S1 post-patching with S2} & \textbf{Align S1 post-patching with S1} \\
        \hline\hline
        Sen12ms & 42.08 & 44.56 & 45.54 \\
        \hline
        Bigearthnet5 & 70.71 & 71.32 & 73.76 \\
        \hline
        Bigearthnet19 & 50.23 & 52.53 & 56.07 \\
        \hline
        Bigearthnet43 & 22.33 & 23.13 & 24.03 \\
        \hline
    \end{tabular}
    }
    \caption{Ablation study results comparing the performance of different alignment strategies for Sentinel-1 data on various RS image classification datasets. The table presents mean Average Precision (mAP) scores. Both aligning Sentinel-1 modality post-patching with either Sentinel-1 or Sentinel-2 data outperforms alignment without patching across all datasets, demonstrating the importance of the patching process for learning robust RS representations, even when patching using data of another RS modality.}
    \label{tab:ablation_s1}
\end{table*}
\endgroup

\begingroup
\setlength{\tabcolsep}{24pt}
\renewcommand{\arraystretch}{1.2}
\begin{table}[h]
    \centering
    \resizebox{\linewidth}{!}{%
    \begin{tabular}{|c||c|c|}
        \hline \textbf{Dataset} & \textbf{RGB Bands} & \textbf{All Bands} \\
        \hline\hline
        EuroSAT & 38.72 & 42.11 \\
        \hline
        SEN12MS & 36.75 & 49.98 \\
        \hline
        BigEarthNet-5 & 70.98 & 79.18 \\
        \hline
        BigEarthNet-19 & 62.47 & 74.69 \\
        \hline
        BigEarthNet-43 & 27.40 & 32.87 \\
        \hline
    \end{tabular}
    }
    \caption{Ablation study to assess whether the Sentinel-2 modality encoder, after cross-modal alignment, learns to utilize the full multi-spectral information or merely reconstructs the RGB composite values from the corresponding bands. The table compares zero-shot classification performance when aligning the encoder using only RGB Bands versus all available Sentinel-2 spectral bands. Both encoders were initialized with the same pre-trained weights; the RGB-only encoder utilized only the weights corresponding to the RGB input channels, while the all-bands encoder utilized all available pre-trained weights. The significantly improved performance when using all spectral bands demonstrates that the encoder learns to leverage information beyond RGB.}
    \label{tab:rgb_vs_ms}
\end{table}
\endgroup

\section{Discussion and Conclusion} \label{sec:discussion}
We present a method aiming to create a joint embedding space for distinct RS imagery modalities. Our proposed method, leverages CLIP large scale pre-training, working towards a RS Vision-Language model (VLM), without relying on textual descriptions, without training from scratch, without introducing any task-specific parameters and without catastrophic forgetting. We enable, in a computationally efficient manner, a rich set of multi-modal tasks across different modalities and set preliminary baselines for cross-modal retrieval and text-based zero-shot tasks using RS imagery. We aspire for our method to serve as a stepping stone for the EO community and facilitate novel applications on RS imagery.

\subsection{A blueprint towards satellite vision-language models}
We view our proposed methodology as a blueprint for the resource-efficient development of a RS Vision-Language model (VLM). Our experimental results highlight two major challenges that warrant further attention: (i) the generalizability of our generated VLM and (ii) the need for a clearer understanding of state-of-the-art VLMs in comparison to well-established standards and benchmarks.

In the pursuit of enhancing the generalizability of our method, it becomes essential to explore specific considerations and refinements that transcend the immediate scope of our experimental setup. We deliberately patch CLIP using a single proxy dataset, as shown in Fig.~\ref{fig:patching_clip}. Nevertheless, it should be obvious that it is extremely difficult for a single dataset to address multiple training objectives (i.e. multi-class vs multi-label classification), as well as the unique challenges posed by RS imagery in terms of spatial coverage, atmospheric conditions, varying perspectives, temporal and spatial resolution, which can significantly impact the performance of DL models. We summarize the characteristics of the datasets used during the course of our experiments in Table~\ref{tab:datasets}. For completeness, we also provide a direct comparison of per dataset sampled image-label pairs in Fig.~\ref{fig:datasets}, where several differences become apparent, such as: i) the spatial resolution differences, ii) the diversity with regards to the level of nomenclature detail used, as well as iii) the effect of atmospheric corrections which in the case of BigEarthNet mitigate atmospheric interference thus enhancing true surface reflectance and reducing color distortion. A straightforward modification to the CLIP model patching method tailored on multiple datasets~\citep{ilharco2022patching}, should remedy the shortcomings observed in our experimental outcomes. Such a modification is expected to significantly improve the performance of patched models, thereby leading also to notable improvements at the cross-modal alignment stage. 

A similar modification could be also introduced to our cross-modal alignment stage, seen in Fig.~\ref{fig:align_clip}. We are confident that such a modification would substantially improve our experimental results summarized in Fig.~\ref{fig:patching_eval} and Table~\ref{tab:clip_retrieval}. Especially for multi-class classification tasks, such as EuroSAT, where our results are currently lacking in terms of performance, since this stems from our current way of aligning the Self-supervised Learning (SSL) pre-trained satellite encoder using solely the BigEarthNet dataset and as a consequence a multi-label optimization objective for the label loss part of our proposed loss function Eq.~\ref{eq:alignment_loss}. Another possible improvement could also be the use of a larger SSL pre-trained satellite modality encoder. As of today, \citet{wang2022ssl4eo} have published a series of pre-trained Sentinel-2 encoder weights, featuring different SSL methods, yet only for ViT-S-16 architecture. This could possibly pose a limiting factor for our cross-modal alignment outcome, since it might outstrip the capacity of our aligned encoder to acquire concepts, especially when the knowledge is transferred from considerably larger CLIP encoders.

Turning our attention to benchmarking, a comprehensive examination of our approach's performance in comparison to established standards is essential to validate its robustness and potential applications in diverse contexts. To this end, recent works~\citep{roberts2023satin, lacoste2023geo, bountos2023fomo} have been focusing on enabling a reliable assessment of progress, in the field of EO foundation models, by fabricating EO benchmark suites alongside robust methodologies for evaluating models and reporting results. Such evaluation suites, along the one used for the evaluation of CLIP~\citep{radford2021learning}, could ultimately provide the community with ranked candidate CLIP models and streamline future application within the EO era.

\subsection{Bridging modalities with distinct intrinsic features}
The cross-modal alignment of complementary RS imagery modalities, beyond RGB, is very important in the context of the EO community. Working towards this end, we pinpoint a pivotal juncture for the successful outcome of our cross-modal alignment stage, specifically the information gap between heterogenetic modalities~\citep{shi2023towards, fu2020mcen}. Prompted by prior works~\citep{liang2022mind}, which determined that contrastive learning preserves the gap between different modalities, we suggest an alternative loss \ref{eq:alignment_loss} formulated as the linear combination of Mean Squared Error (MSE) and Cross Entropy (CE) losses. Contrary to \citet{girdhar2023imagebind}, which trains a joint embedding space for multiple modalities using solely image alignment, we leverage guidance from the scaled CE signal in an effort to alleviate the modality gap of RGB composites and RS imagery and allow the RS modality encoder to better align within the shared embedding space while being constrained to produce similar image embedddings, due to the MSE loss. All in all, our alignment procedure resembles a plain cross-modal knowledge distillation setup, where the knowledge distillation (KD) loss has been altered to imitate the output embeddings of the teacher network, in contrast to the softened class scores. CLIP knowledge distillation has been receiving a lot of attention lately and has already proven its capabilities on a series of works~\citep{wu2023tinyclip, yang2023clipkd, pei2023clipping, wang2022clip, wu2022wav2clip, guzhov2022audioclip}, which aspire to exploit the large scale pre-training of CLIP in an computationally efficient and interoperable way.

\section*{Acknowledgment}
This work has received funding from the European Union's \\ Horizon Europe research and innovation project ThinkingEarth under grant agreement number 101130544.

\appendix

\section{Patched models zero-shot performance} 
\label{ap:patchedmodelszeroshot}
In this section we report in detail the patching results for all four CLIP ViT-based models for the patching task (i.e. BigEarthNet-19) and the representative supported task (i.e. ImageNet) for the different $\alpha$ weight interpolation coefficients. For completeness, we also include the zero-shot performance evaluation results of the patched models, for the different $\alpha$ coefficients, on a set of diverse satellite scene classification nomenclatures and datasets. See Table~\ref{tab:patching_metrics}.

\begingroup
\setlength{\tabcolsep}{21pt} 
\renewcommand{\arraystretch}{1.2} 
\begin{table*}[b]
    \centering
    \resizebox{\textwidth}{!}{\begin{tabular}{|c|c|c|c|c|c|c|c|c|c|c|c|}
        \hline \multirow{2}{*}{ \textbf{ViT-B-32} } & \multicolumn{11}{c|}{ $\pmb{\alpha}$ } \\
        \cline{2-12} 
        & \textbf{0,00} & \textbf{0,10} & \textbf{0,20} & \textbf{0,30} & \textbf{0,40} & \textbf{0,50} & \textbf{0,60} & \textbf{0,70} & \textbf{0,80} & \textbf{0,90} & \textbf{1,00} \\
        \hline BigEarthNet-19 (mAP) & 15,18 & 15,18 & 15,18 & 15,86 & 28,66 & 51,13 & 67,79 & 73,37 & 75,52 & 76,46 & 76,76 \\
        \hline ImageNet (acc) & 63,28 & 63,19 & 62,67 & 61,97 & 60,71 & 58,65 & 55,04 & 49,45 & 40,95 & 30,57 & 19,74 \\
        \hline EuroSAT (acc) & 49,98 & 56,93 & 60,70 & 58,11 & 57,13 & 56,57 & 54,07 & 51,28 & 48,31 & 45,78 & 43,87 \\
        \hline SEN12MS (mAP) & 22,83 & 22,83 & 22,83 & 22,83 & 26,04 & 40,93 & 46,98 & 48,15 & 47,84 & 47,31 & 46,80 \\
        \hline BigEarthNet-5 (mAP) & 35,16 & 35,16 & 35,16 & 36,93 & 50,57 & 60,30 & 68,12 & 70,52 & 71,37 & 72,09 & 72,80 \\
        \hline BigEarthNet-43 (mAP) & 6,82 & 6,82 & 6,82 & 7,27 & 12,89 & 21,97 & 29,27 & 32,12 & 32,84 & 33,05 & 33,01 \\
        \hline \hline \multirow{2}{*}{ \textbf{ViT-B-16} } & \multicolumn{11}{c|}{ $\pmb{\alpha}$ } \\
        \cline{2-12}
        & \textbf{0,00} & \textbf{0,10} & \textbf{0,20} & \textbf{0,30} & \textbf{0,40} & \textbf{0,50} & \textbf{0,60} & \textbf{0,70} & \textbf{0,80} & \textbf{0,90} & \textbf{1,00} \\
        \hline BigEarthNet-19 (mAP) & 15,18 & 15,18 & 15,18 & 15,28 & 20,98 & 39,36 & 63,64 & 72,61 & 75,58 & 76,92 & 77,46 \\
        \hline ImageNet (acc) & 68,33 & 68,36 & 68,05 & 67,40 & 66,07 & 63,59 & 59,05 & 51,74 & 41,14 & 28,78 & 17,32 \\
        \hline EuroSAT (acc) & 55,02 & 52,44 & 48,20 & 43,70 & 39,61 & 35,50 & 34,28 & 35,59 & 35,72 & 34,06 & 31,69 \\
        \hline SEN12MS (mAP) & 22,83 & 22,83 & 22,83 & 22,83 & 23,10 & 28,65 & 38,86 & 45,78 & 45,94 & 44,61 & 43,27 \\
        \hline BigEarthNet-5 (mAP) & 35,16 & 35,16 & 35,16 & 35,40 & 44,70 & 59,92 & 71,56 & 76,41 & 76,10 & 75,44 & 75,00 \\
        \hline BigEarthNet-43 (mAP) & 6,82 & 6,82 & 6,82 & 6,93 & 9,43 & 16,75 & 27,09 & 31,39 & 32,11 & 32,36 & 32,48 \\
        \hline \hline \multirow{2}{*}{ \textbf{ViT-L-14} } & \multicolumn{11}{c|}{ $\pmb{\alpha}$ } \\
        \cline{2-12} 
        & \textbf{0,00} & \textbf{0,10} & \textbf{0,20} & \textbf{0,30} & \textbf{0,40} & \textbf{0,50} & \textbf{0,60} & \textbf{0,70} & \textbf{0,80} & \textbf{0,90} & \textbf{1,00} \\
        \hline BigEarthNet-19 (mAP) & 15,18 & 15,20 & 15,85 & 28,91 & 56,10 & 72,04 & 78,04 & 80,72 & 81,30 & 81,63 & 81,53 \\
        \hline ImageNet (acc) & 75,44 & 75,32 & 75,19 & 74,98 & 75,32 & 74,22 & 73,57 & 72,62 & 71,25 & 69,00 & 65,72 \\
        \hline EuroSAT (acc) & 62,48 & 67,50 & 69,54 & 70,41 & 69,50 & 69,70 & 68,28 & 66,87 & 64,93 & 60,76 & 59,70 \\
        \hline SEN12MS (mAP) & 22,83 & 22,83 & 22,84 & 24,65 & 34,25 & 43,44 & 47,31 & 48,52 & 48,13 & 47,78 & 47,58 \\
        \hline BigEarthNet-5 (mAP) & 35,16 & 35,16 & 35,94 & 55,02 & 67,45 & 74,92 & 77,56 & 77,55 & 77,54 & 77,73 & 75,83 \\
        \hline BigEarthNet-43 (mAP) & 6,82 & 6,83 & 7,05 & 12,88 & 25,25 & 35,26 & 39,80 & 40,00 & 40,86 & 39,64 & 39,08 \\
        \hline \hline \multirow{2}{*}{ \textbf{ViT-L-14-336} } & \multicolumn{11}{c|}{ $\pmb{\alpha}$ } \\
        \cline{2-12} 
        & \textbf{0,00} & \textbf{0,10} & \textbf{0,20} & \textbf{0,30} & \textbf{0,40} & \textbf{0,50} & \textbf{0,60} & \textbf{0,70} & \textbf{0,80} & \textbf{0,90} & \textbf{1,00} \\
        \hline BigEarthNet-19 (mAP) & 15,18 & 15,18 & 15,85 & 28,91 & 60,43 & 73,53 & 78,05 & 80,09 & 81,12 & 81,63 & 81,86 \\
        \hline ImageNet (acc) & 76,56 & 76,53 & 76,41 & 76,21 & 75,92 & 75,53 & 74,97 & 74,04 & 72,67 & 70,70 & 67,29 \\
        \hline EuroSAT (acc) & 61,70 & 67,96 & 69,26 & 70,41 & 69,30 & 69,24 & 68,20 & 66,26 & 63,89 & 60,76 & 57,98 \\
        \hline SEN12MS (mAP) & 22,83 & 22,83 & 22,83 & 22,84 & 28,79 & 40,78 & 46,85 & 48,52 & 48,13 & 47,78 & 47,58 \\
        \hline BigEarthNet-5 (mAP) & 35,16 & 35,16 & 35,27 & 55,02 & 71,10 & 77,28 & 77,73 & 77,55 & 77,54 & 77,73 & 78,02 \\
        \hline BigEarthNet-43 (mAP) & 6,82 & 6,82 & 6,92 & 13,31 & 27,54 & 37,03 & 39,80 & 40,72 & 39,50 & 40,55 & 40,05 \\
        \hline
    \end{tabular}}
    \caption{Zero-shot performance of patched CLIP~\citep{radford2021learning} ViT models on a set of diverse satellite scene classification nomenclatures and datasets. We consider BigEarthNet-19 as a proxy dataset for the satellite scene classification task, and ImageNet as a representative supported task. The table compares all four CLIP ViT architectures (ViT-B-32, ViT-B-16, ViT-L-14, ViT-L-14-336) using different weight interpolation values $\alpha \in [0.0,1,0]$. Performance is measured in terms of either Accuracy (acc) or mean Average Precision (mAP) on six datasets: Imagenet (acc), BigEarthNet-5 (mAP), BigEarthNet-19 (mAP), BigEarthNet-43 (mAP), EuroSAT (acc), and SEN12MS (mAP). Results demonstrate how model size and $\alpha$ values affect classification performance across diverse satellite imagery tasks.}
    \label{tab:patching_metrics}
\end{table*}
\endgroup

\section{Per-class performance}
\label{ap:bestmodel}
\begin{figure}[t]
    \centering
    \includegraphics[width=1.0\linewidth]{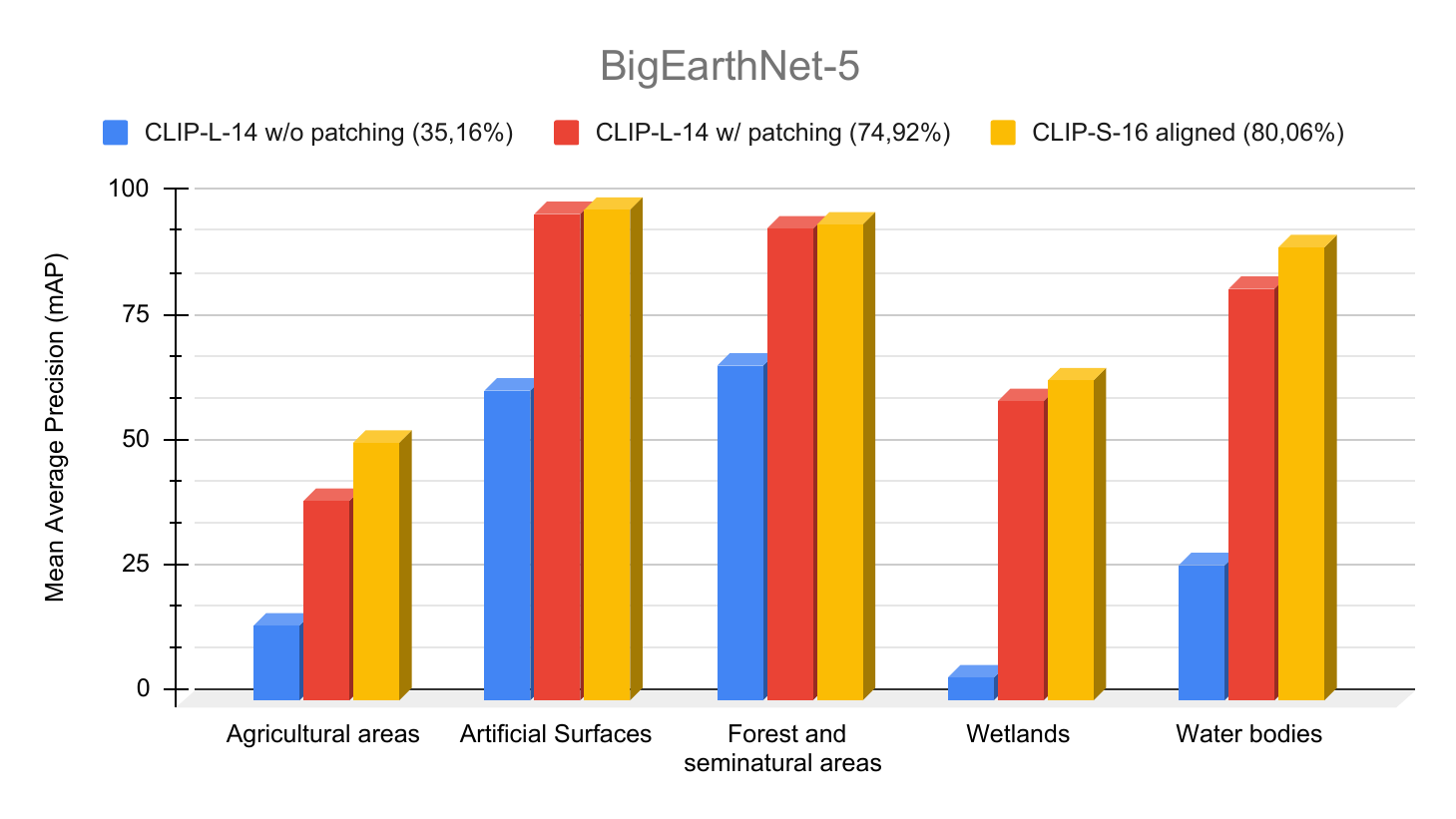}
    \caption{Zero-shot performance comparison of our best performing CLIP-L-14 model on BigEarthNet-43 RGB, pre- and post-patching on BigEarthNet-19 RGB. Additionally, performance of the pre-trained ViT-S-16 after alignment with the patched CLIP-L-14 model. BigEarthNet-5 features a much simpler 5-class nomenclature, in comparison to the BigEarthNet-19 19-class one. Results indicate significant improvements in zero-shot classification mean Average Precision (mAP) through patching and alignment stages for the task of satellite scene classification.}
    \label{fig:be5_rgb}
\end{figure}

\begin{figure*}
    \centering
    \includegraphics[width=1.0\linewidth]{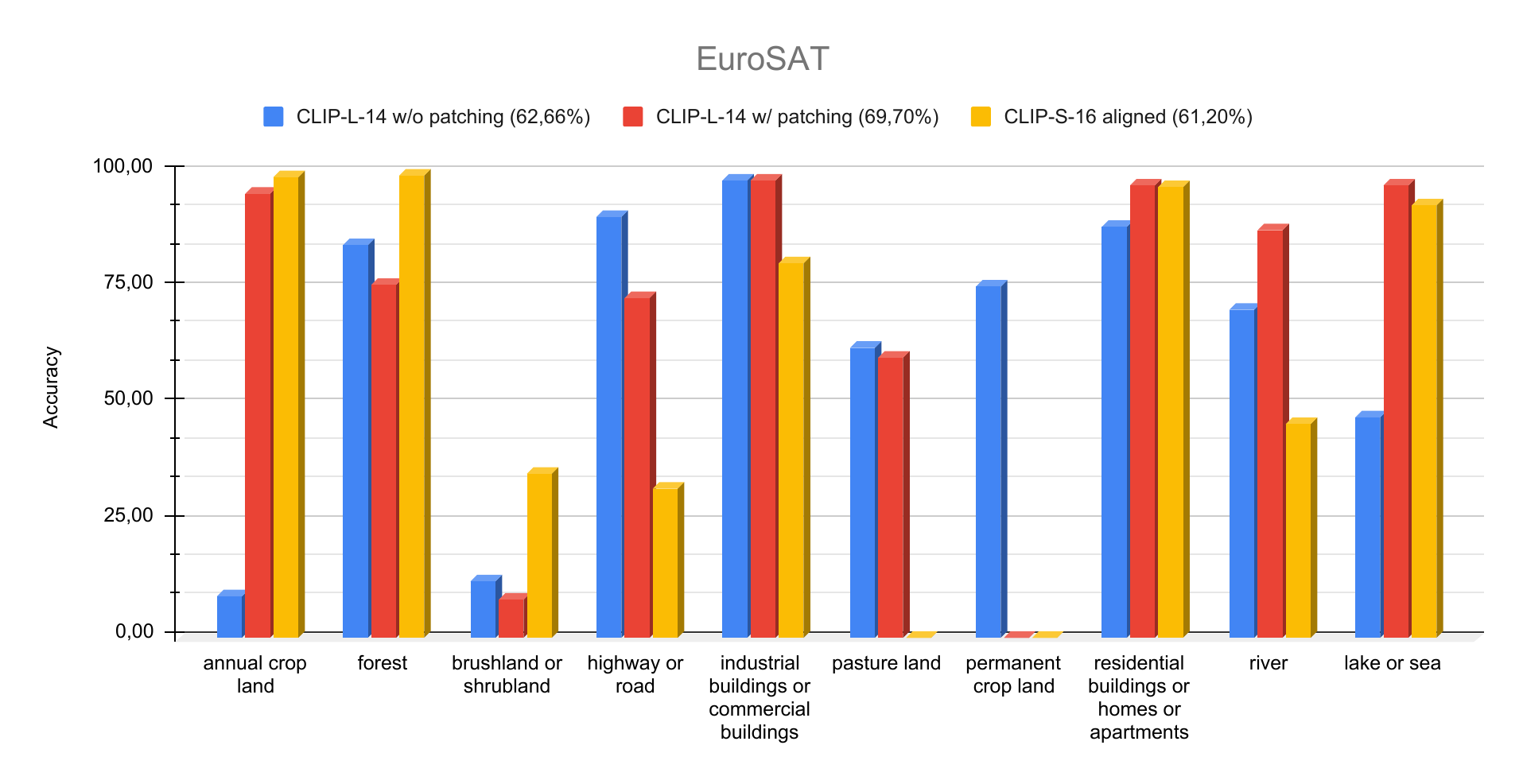}
    \caption{Zero-shot performance comparison of our best performing CLIP-L-14 model on EuroSAT RGB, pre- and post-patching on BigEarthNet-19. Additionally, performance of the pre-trained ViT-S-16 after alignment with the 
    patched CLIP-L-14 model. EuroSAT features a 10-class nomenclature. Results indicate significant improvements in zero-shot classification Accuracy through patching and alignment stages for the task of satellite scene classification.}
    \label{fig:eurosat_rgb}
\end{figure*}

\begin{figure*}
    \centering
    \includegraphics[width=0.95\linewidth]{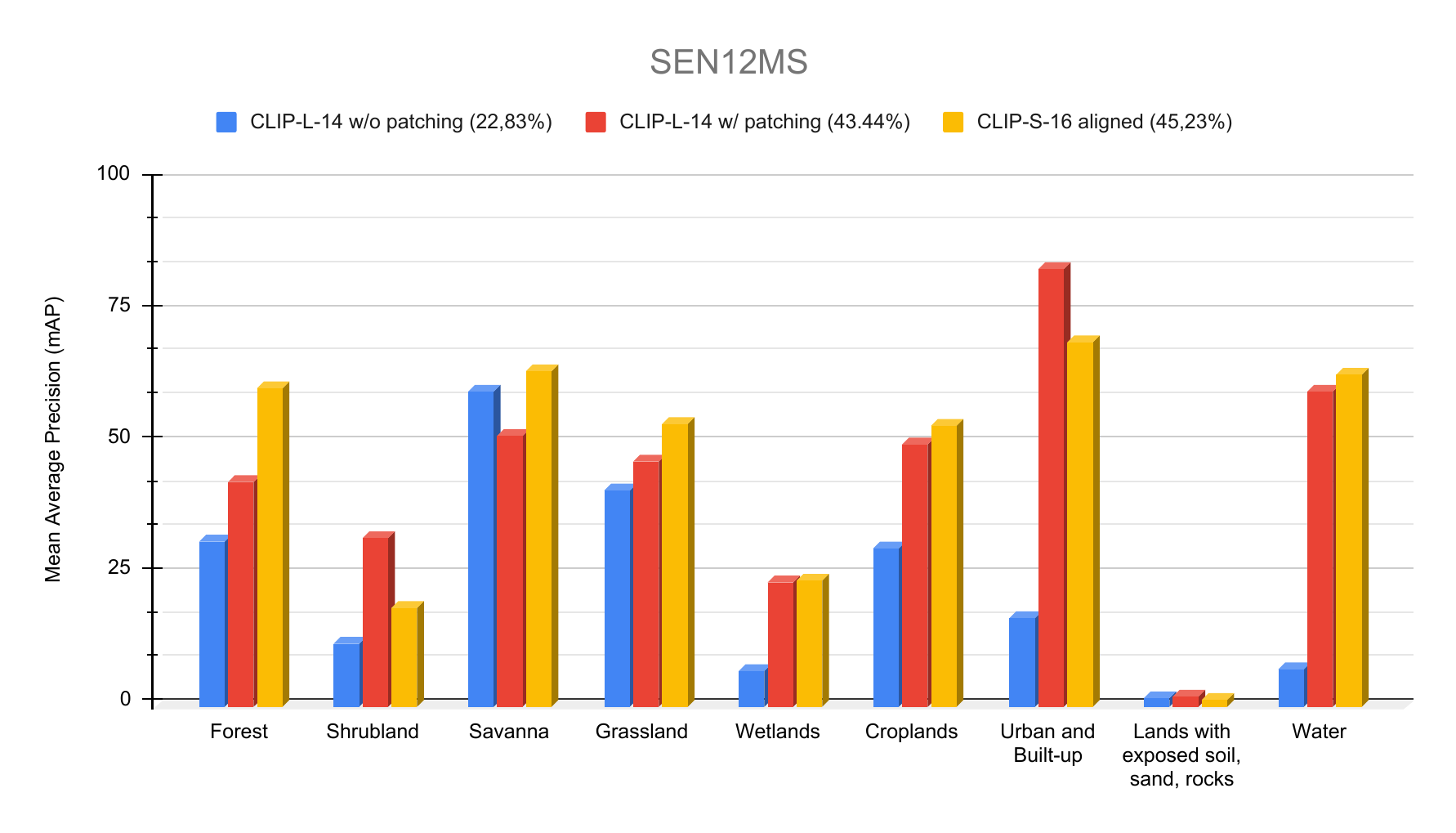}
    \caption{Zero-shot performance comparison of our best performing CLIP-L-14 model on SEN12MS RGB, pre- and post-patching on BigEarthNet-19 RGB. Additionally, performance of the pre-trained ViT-S-16 after alignment with the 
    patched CLIP-L-14 model. SEN12MS features a 9-class nomenclature. Results indicate significant improvements in zero-shot classification mean Average Precision (mAP) through patching and alignment stages for the task of satellite scene classification.}
    \label{fig:sen12ms_rgb}
\end{figure*}

\begin{figure*}
    \centering
    \includegraphics[width=1.0\linewidth]{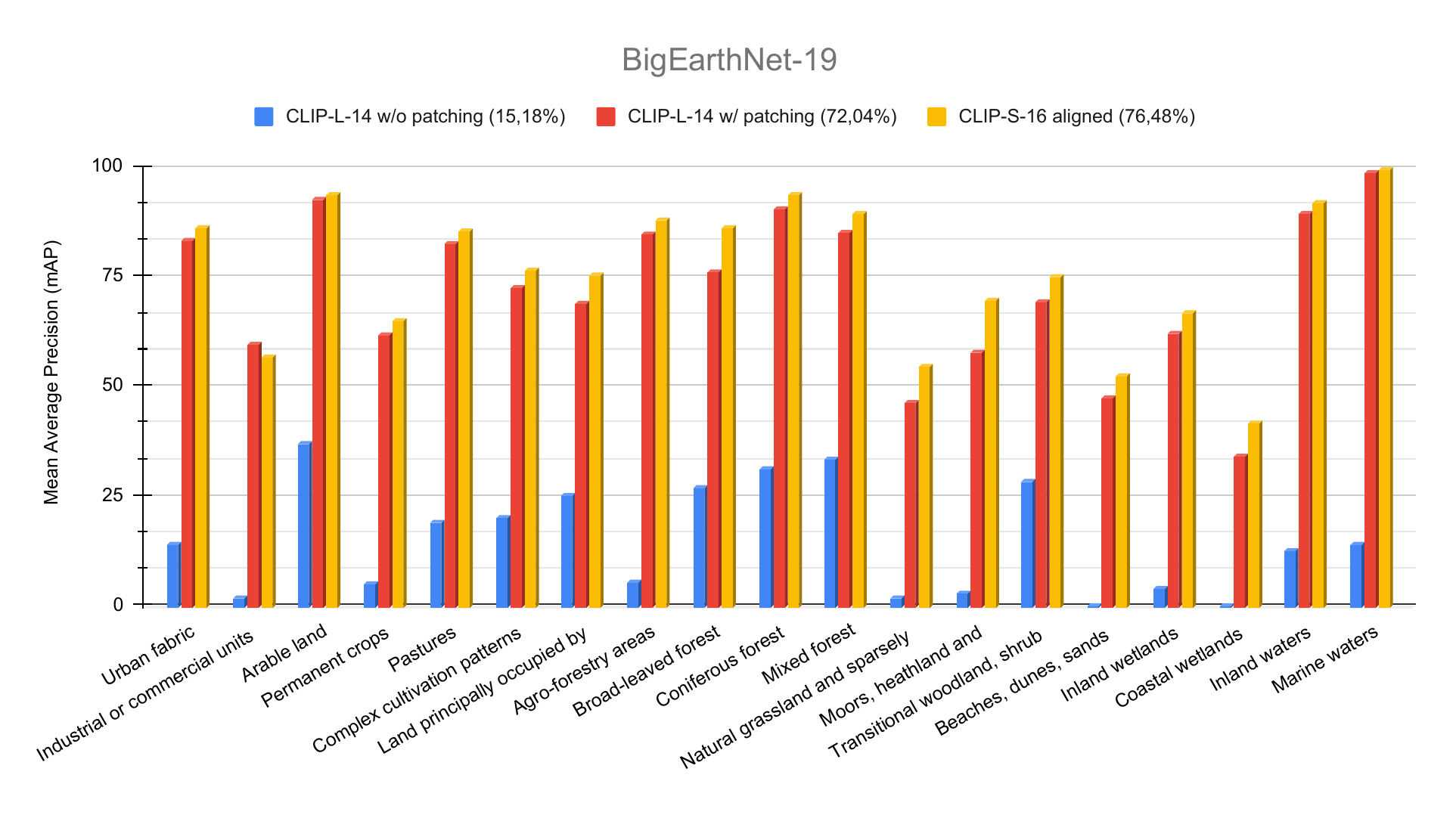}
    \caption{Performance comparison of our best performing CLIP-L-14 model, pre- and post-patching on BigEarthNet-19 RGB. Additionally, performance of the pre-trained ViT-S-16 after alignment with the patched CLIP-L-14 model. Results indicate significant improvements in zero-shot classification mean Average Precision (mAP) through patching and alignment stages for the task of satellite scene classification.}
    \label{fig:be19_rgb}
\end{figure*}

\begin{figure*}[t]
    \centering
    \includegraphics[width=1\linewidth]{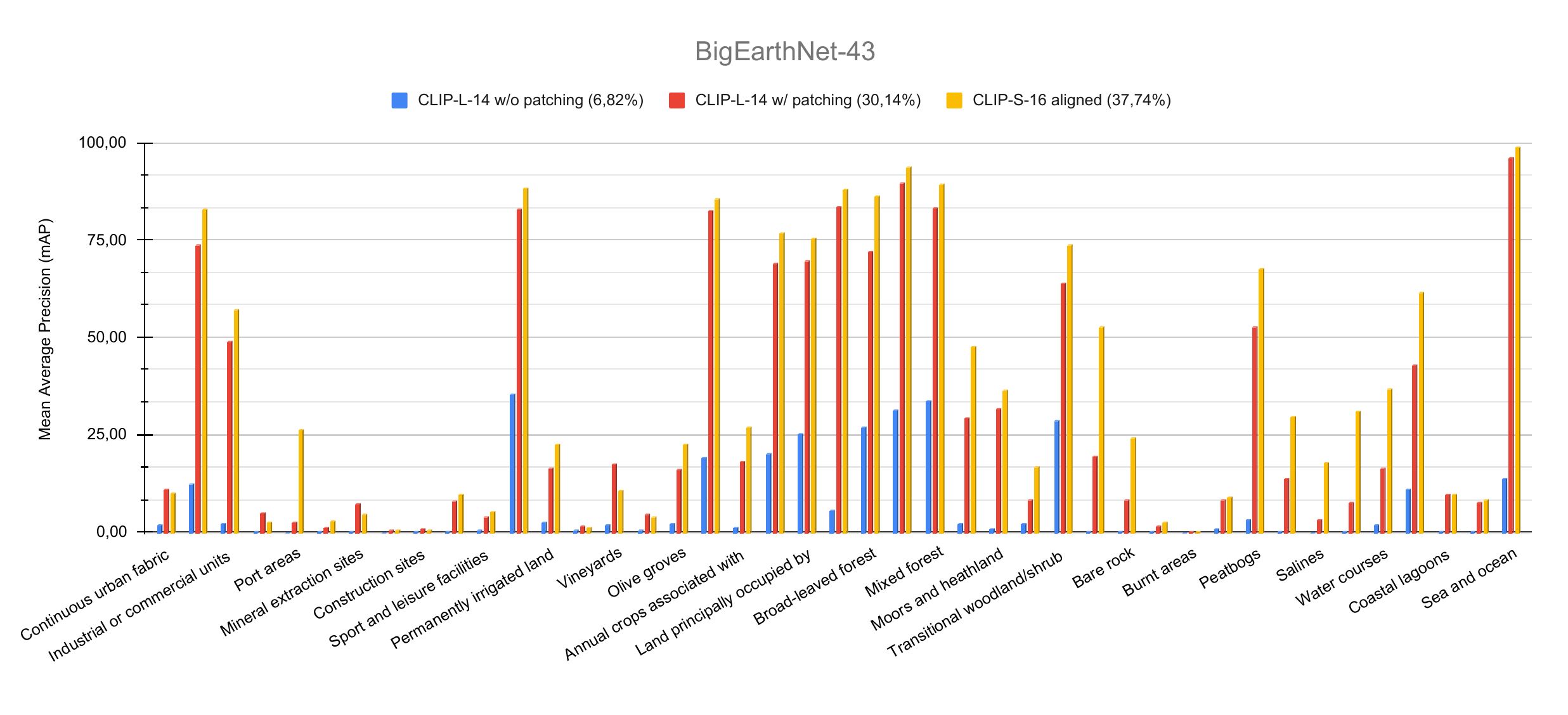}
    \caption{Zero-shot performance comparison of our best performing CLIP-L-14 model on BigEarthNet-43 RGB, pre- and post-patching on BigEarthNet-19 RGB. Additionally, performance of the pre-trained ViT-S-16 after alignment with the patched CLIP-L-14 model. BigEarthNet-43 features a detailed 43-class nomenclature, in comparison to the BigEarthNet-19 19-class one. Results indicate significant improvements in zero-shot classification mean Average Precision (mAP) through patching and alignment stages for the task of satellite scene classification.}
    \label{fig:be43_rgb}
\end{figure*}

In this section we present the per-class performance of our best CLIP-L-14 model. As always we present the performance of the model pre- and post- patching with BigEarthNet-19. Additionally the plots contain the performance of the ViT-S-16 after alignment with the CLIP-L-14 model. Figures~\ref{fig:be5_rgb} and~\ref{fig:be19_rgb} present per-class performance on BigEarthNet-5 and BigEarthNet-19 respectively, while Figure~\ref{fig:be43_rgb} present per-class performance on BigEarthNet-43. Finally, Figures~\ref{fig:sen12ms_rgb} and~\ref{fig:eurosat_rgb} contain per-class performance on SEN12MS and EuroSAT. 

\cleardoublepage\clearpage

\bibliographystyle{elsarticle-harv}
\bibliography{main}

\end{document}